%% file: main.tex
\documentclass[twoside]{article}

\usepackage[accepted]{aistats2023}
\usepackage[round]{natbib}
\usepackage[utf8]{inputenc} 
\usepackage[T1]{fontenc}    
\usepackage{hyperref}       
\usepackage{url}            
\usepackage{booktabs}       
\usepackage{amsfonts}       
\usepackage{nicefrac}       
\usepackage{microtype}      
\usepackage{xcolor}         
\usepackage{amsthm}
\usepackage{amsmath}
\usepackage{amssymb }
\usepackage{algorithm}
\usepackage[noend]{algorithmic}
\usepackage{diagbox}
\usepackage{makecell}
\usepackage{graphicx}
\usepackage{tcolorbox}
\usepackage[para]{footmisc}
\DeclareMathOperator*{\argmax}{arg\,max}

\renewcommand{\vec}[1]{\mathbf{#1}}
\makeatletter
\newcommand\footnoteref[1]{\protected@xdef\@thefnmark{\ref{#1}}\@footnotemark}

\begin{document}

\twocolumn[

\aistatstitle{Bayesian Optimization Over Iterative Learners with Structured Responses: A Budget-aware Planning Approach}

\aistatsauthor{Syrine Belakaria$^+$ \And Janardhan Rao Doppa$^+$ \And  Nicolo Fusi$^\dagger$ \And Rishit Sheth$^\dagger$}

\aistatsaddress{Washington State University$^+$ \And  Microsoft Research$^\dagger$ } ]

\begin{abstract}
The rising growth of deep neural networks (DNNs) and datasets in size motivates the need for efficient solutions for simultaneous
model selection and training.
Many methods for hyperparameter optimization (HPO) of iterative learners, including DNNs, attempt to
solve this problem by querying and learning a response surface while searching for the
optimum of that surface.
However, many of these methods make myopic queries, do not consider
prior knowledge about the response structure, and/or perform a biased cost-aware search, all
of which exacerbate identifying the best-performing model when a total cost
budget is specified. This paper proposes a novel approach referred to as {\bf B}udget-{\bf A}ware {\bf P}lanning for {\bf I}terative Learners (BAPI) to solve HPO problems under a constrained cost budget. BAPI is an efficient non-myopic Bayesian optimization solution that accounts for the budget and leverages the prior knowledge about the objective function and cost function to select better configurations and to take more informed decisions during the evaluation (training). Experiments on diverse HPO benchmarks for iterative learners show that BAPI performs better than state-of-the-art baselines in most cases. 
\end{abstract}

\input{introduction.tex}
\input{Background_motivation}
\input{proposed_approach}
\input{Related_work}
\input{experiments}
\section{SUMMARY}
This paper considered the problem of hyperparameter optimization (HPO) for iterative learners under a constrained cost budget. The proposed BAPI approach addressed gaps in prior work including modeling of structured responses and mis-calibration between response and cost models leading to biased search. More importantly, our planning-based BAPI approach allows for non-myopic candidate selection over horizons adaptive to the budget. Combined with subset selection and early termination procedures, our experimental evaluation on a variety of HPO benchmarks shows BAPI's efficacy over previous methods in finding high-performing candidates with less cost budget. 
\newpage

\input{main.bbl}
\appendix
\onecolumn
\aistatstitle{Bayesian Optimization Over Iterative Learners with Structured Responses: 
A Budget-aware Planning Approach: \\
Supplementary Materials}

\input{appendix.tex}

\end{document}

%% file: introduction.tex
\section{INTRODUCTION}\label{intro}
Hyperparameter optimization (HPO) for machine learning models, and pipelines is the task of automatic tuning of those parameters which affects model selection and training.
A variety of HPO approaches have been developed for classical ML models, e.g., SVMs, random forests, utilizing
Bayesian optimization  (BO) \citep{snoek2012practical,swersky2013multi},
meta-learning \citep{feurer-neurips15a,fusi2018probabilistic},
and ensembling \citep{thornton2013auto,olson2016tpot,ledell2020h2o,erickson2020autogluon}
to name a few.
For HPO, the inclusion of modern deep neural networks (DNNs) as a pipeline, introduces a temporal dimension to the problem of selecting  pipeline queries due to the iterative nature of DNNs. Fixing the number of training epochs per query is clearly inefficient since poorly performing pipelines will waste resources,
leading to more promising candidates not being identified when the resource budget is small. 
This paper considers the problem of HPO for such iterative learners (ILs) under a {\em fixed total budget}. In this setting, the budget is defined as some measure of resource cost for evaluating queries such as wallclock time or energy. The key challenge is to reason about the available budget to intelligently select the candidate queries for evaluation to uncover high-quality configurations within the remaining budget. BO is known to be an effective framework to solve such problems. However, most BO algorithms ignore the fact that the cost of different configurations/queries can vary significantly and are unknown prior to their evaluation. We refer to this problem of utilizing BO to select amongst iterative learners under a constrained budget as \emph{budget-aware Bayesian optimization}.

The key idea behind BO for HPO is to learn a response surface (e.g., Gaussian process) which serves as a surrogate for test set performance and use it to perform a sequence of queries by trading off exploration and exploitation. The response and cost modeling, and the planning of querying a sequence of different configurations have individually been addressed in the BO literature. To some extent, the interaction of these two components was studied as well. For ILs, modeling the side-information in the form of the shape of learning curves (i.e., accuracy vs. training epochs) will allow us to make fine-grained decisions such as early stopping to save resources. In our problem setting, we refer to the learning curve as the {\em structured response}. The structure of responses have been considered and modeled to varying
degrees in methods that extrapolate performance to determine good candidates \citep{klein2016learning,domhan2015speeding} as well as in BO over vector-valued responses \citep{wu2020practical,nguyen2020bayesian}. However, none of these settings consider a fixed total budget. 

To summarize the drawbacks of prior work for our problem setting: standard approaches for handling structured responses, heterogeneous cost-modeling (i.e., different queries have varying costs), and query selection have inefficiencies or are even inaccurate in some cases, especially when applied to the fixed budget setting. Standard cost-aware modeling for BO utilizes a separate model for cost predictions and
then weights the selection of the next candidate query by these predictions \citep{snoek2012practical}.
However, cost-aware modeling via weighting suffers from a known pathology where the method tends to select low-cost queries with lower accuracy leading to an overall poor performance \citep{lee2020cost,astudillo2021multi}. Fundamentally, the issue is one of mis-calibration between the response and cost models.

For query selection, non-myopic BO methods provide approximations of varying quality to the optimal solution
defined by the Bellman recursion for BO \citep{osborne2010bayesian,lam2016bayesian,jiang2020binoculars}.
These non-myopic methods \citep{lam2016bayesian,wu2019practical,yue2020non,jiang2020efficient,lee2020efficient,lee2021nonmyopic,astudillo2021multi}
use a variety of techniques to solve the sequence of nested integrations and maximizations 
including dynamic programming and rollouts.
However, only the two most recent methods of \citet{lee2021nonmyopic} and \citet{astudillo2021multi} take the cost and a finite budget into account.
\citet{lee2021nonmyopic} proposed to leverage the known pathology of standard cost-aware modeling to promote early
exploration but the resulting policy {\em does not adapt its horizon to the remaining budget}.
\citet{astudillo2021multi} also propose a non-myopic policy, but the {\em horizon adaptation of their method is post-hoc}, i.e. zero padding is used to fill the horizon after the evaluation budget is exhausted.

\noindent {\bf Contributions}\hspace{10pt} Our proposed solution BAPI executes a non-myopic query selection policy by wrapping standard BO in a layer of budget-aware planning for iterative learners. The key innovations of BAPI include leveraging side-information and expert knowledge such as the objective function's monotonicity, heteorgenity of query costs, and linearity of cost w.r.t training epochs into the planning procedure; overcoming known pathologies of standard cost-aware BO; and principled approach for adapting the horizon to the amount of remaining budget.
Therefore, our technical contributions span both budgeted non-myopic BO and hyper-parameter optimization sub-areas.
The list of synergistic contributions made by this paper are as follows:
\vspace{-6mm}
\begin{itemize}
\setlength\itemsep{0em}
    \item Development of a new approach for budget-aware non-myopic BO enabling an adaptive horizon to solve HPO problems for iterative learners. To the best of our knowledge, this is the first work that proposes a budgeted non-myopic approach specifically for HPO.
    \item Refining previous response modeling approaches by leveraging the monotonicity of the objective function through model derivatives to enable: (1) a new conservative stopping estimation approach to decide when a learner becomes $\epsilon$-close to its asymptotic value, and (2) a general modeling approach with minimal assumptions about the shape of the response resulting in accurate extrapolation for improved decision-making.
    \item Design of a new efficient early termination method aimed  to early stop the training of poorly performing HP configurations.
    \item A new alternative kernel for modeling the training cost of iterative learners while capturing the linearity of the cost w.r.t the number of epochs and its variability across different HPs
    \item Empirical evaluation on several state-of-the art benchmarks to demonstrate the performance of BAPI compared to algorithms designed for HPO, generic BO, and non-myopic BO and cost-aware non-myopic BO. 
\end{itemize}

%% file: Background_motivation.tex
\section{PROBLEM SETUP AND BACKGROUND}\label{background}
In this section, we state our problem and briefly review Gaussian processes, Bayesian optimization, and myopic vs.
non-myopic query selection policies.

Consider the problem of sequentially optimizing a blackbox objective function $f$ over the input space $\mathcal{X}$ where the evaluation of each candidate input $\vec{x} \in \mathcal{X}$ is expensive and where the cost $c$ of each input is unknown before the evaluation. In the context of HPO for iterative machine learning models, each input candidate $\vec{z}:=[\vec{x},t]$, where $\vec{x}$ represents model/pipeline hyperparameters and $t\in \mathcal{T}$=$[1\dots t_{max}]$ is the number of training epochs.
We let the objective function $f(\vec{x},t)$ be defined as the accuracy \footnote{Any bounded metric can be used (e.g, loss, some cases of reward for RL models etc.)} and the unknown cost $c(\vec{x},t)$ be defined as the training time.
The objective is to identify the maximum of $f$ in a number of queries whose cumulative cost is bounded by a total budget
$B_T$. 
Let $\mathcal{Z}:=\mathcal{X} \times \mathcal{T}$, our problem can be stated as
\vspace{-1mm}
\begin{equation}
    \max_{Z\in P(\mathcal{Z})} \max_{\vec{z}\in Z} f(\vec{z}), \ \ \ \ \text{~s.t.~} \sum_{\vec{z}\in Z} c(\vec{z}) \le B_T \label{eq:problem}
\end{equation}
where $P(\mathcal{Z})$ denotes the power set of $\mathcal{Z}$ and $Z=\{\vec{z}_1 \dots \vec{z}_h\}$ is the sequence of inputs evaluated until the budget $B_T$ is exhausted.
In other words, the optimal HP $ \vec{z}^*$ is defined as  $ \vec{z}^* \leftarrow \argmax_{\mathcal{Z}}f(\vec{z})$ with $\vec{\vec{z}}^*\in Z$ and $\sum_{\vec{z}\in Z} c(\vec{\vec{z}}) \leq B_T$. The problem in Equation 
(\ref{eq:problem}) is solved using a non-myopic policy, where at each iteration, the algorithm accounts for the sequence of inputs that can be evaluated within the remaining budget, i.e., the horizon $h$ is {\em adaptive}. We define the non-myopic setting later in this section, which is similar to the setting considered in \citet{lee2021nonmyopic} and \citet{astudillo2021multi}.

We focus on problem settings where the objective function is monotonic in the number of epochs $t$.
Specifically, for a fixed hyperparameter $\vec{x}$, $f(\vec{x},t) \leq f(\vec{x},t') \text{~when~}  t\le t'$. 
This is a reasonable assumption for iterative learners. Even if monotonicity does not hold over all training epochs,
keeping track of the best model over training epochs is a standard practice \citep{dai2019bayesian}.

\textbf{Gaussian Processes}\hspace{10pt} GPs are characterized by a mean function $m$ and a covariance or kernel function $K$. If a function $f$ is sampled from GP($m$, $K$), then $f(\vec{x},t)$ is distributed normally $\mathcal{N}(m(\vec{x},t), K([\vec{x},t],[\vec{x},t]))$ for a finite set of inputs from $[\vec{x},t] \in \mathcal{X} \times \mathcal{T}$.
The predictive mean and uncertainty for a GP for a new input $\vec{z}_* \in \mathcal{Z}$
is defined as:
\begin{align}
        &\mu(\vec{z}_*) =  K_{\vec{z}_*,Z}[K_{Z,Z} + \sigma^{2}I]^{-1}(Y - m(Z))+m(\vec{z}_*) \nonumber \\  & \sigma^2(\vec{z}_*)   = K_{\vec{z}_*, \vec{z}_*} - K_{\vec{z}_*, Z}[K_{Z, Z} + \sigma^{2}I]^{-1}K_{Z, \vec{z}_*} \nonumber
\end{align}
where $K_{\vec{z}_*, \vec{z}_*} = K(\vec{z}_*,\vec{z}_*)$,  $K_{Z,Z}=K(Z,Z)$, $K_{\vec{z}_*,Z} = [K(\vec{z}_*,\vec{z}_i)]_{\forall i}$, $Z$ is the set of evaluated inputs and $Y$ is their corresponding function values. A typical choice to model blackbox functions with a temporal component is using a product kernel
$
    K([\vec{x},t],[\vec{x'},t'])=K_{\vec{x}}(\vec{x},\vec{x'})\times K_t(t,t').
$
$K_{\vec{x}}$, defined over the input space $\vec{x}\in \mathcal{X}$, is often selected to be an
RBF 
or a Matern kernel. For the temporal component $K_t$, previous work for GPs over iterative learners \citep{swersky2014freeze} proposed an exponential decay (ED) kernel, defined as  $K_t(t,t')=\beta^\alpha / (t+t'+\beta)^\alpha$, to model decreasing covariance with increasing time. However, this kernel does not guarantee that the predictive mean of GP or sampled functions would necessarily follow a desired monotonic shape. \citet{nguyen2020bayesian} argued that the use of ED is not appropriate for reinforcement learning models where the reward might follow a logistic shape 
and proposed the use of an RBF kernel for $K_t$.

\textbf{Bayesian Optimization And Non-Myopic Query Policies}\hspace{10pt} BO is a sequential, model-based approach for optimizing blackbox functions \citep{shahriari2015taking,MESMO,HyBO}.
BO is often performed with the specification of a GP prior over the function,
and an acquisition function.
The GP posterior acts as a surrogate for the true unknown response. The potential or utility of points in the input space to be
the optimizer is scored by the acquisition function.
Two examples of acquisition functions are expected improvement (EI) and upper confidence bound (UCB) and both are considered myopic since they only aim to maximize the function for the next query without accounting for the future queries. 

We review some standard facts for optimal sequential decision-making \citep{osborne2010bayesian,jiang2020binoculars}.
Consider having collected a set of $i$ responses $D_i$ and let $u$ denote the utility of $D_i$ for maximizing $f(\vec{z})=y$, i.e., 
$ u(D_i)= \max_{(\vec{z},y)\in D_i} y.$
The marginal gain in utility of the query $\vec{z}$ w.r.t.\ $D_i$ is expressed as:
\begin{equation}
    u(y|\vec{z},D_i)= u (D_i \cup { (\vec{z},y)})- u(D_i)
\end{equation}
The one-step expected marginal gain is equivalent to the expected improvement 
(EI) strategy \citep{movckus1975bayesian}:
\begin{equation}
   \mathcal{U}_1(\vec{z}|D_i)= \mathbb{E}_{y}[u (D_i \cup { (\vec{z},y)})- u(D_i) |\vec{z},D_i]
\end{equation}
Now, consider the case where $r$ steps are remaining.
The $r$-steps expected marginal gain can be expressed through the Bellman recursion as \citep{jiang2020binoculars}:
\begin{equation}
   \mathcal{U}_r(\vec{z}|D_i)= \mathbb{E}_{y}[u(y|\vec{z},D_i)]+ \mathbb{E}_{y}[\max_{\vec{z}'}\mathcal{U}_{r-1}(\vec{z}' | D_i \cup {(\vec{z},y)})] \label{eq:1}
\end{equation}
Maximizing (\ref{eq:1}) w.r.t.\ $\vec{z}$ results in the optimal $r$-steps ``lookahead'' .
Being a sequence of $r$ nested integrals of maximizations, optimizing (\ref{eq:1}) is 
intractable for even
small $r$.
\textbf{Lower Bound To Optimal Policy}\hspace{10pt}
The previous discussion focused on the optimality of selecting single queries.
We review recent work by \citep{jiang2020binoculars} which makes a connection between single selection and batch selection of size $r$, $Z =\{\vec{z}_1 \dots \vec{z}_r\}$.
Assuming parallel evaluation, the optimal set of selected points $Z^*$ maximizes the expected marginal utility of the new associated evaluations $Y=\{y_1\dots y_r\}$:
\begin{align}
   &Z^*=\argmax_{Z\in \mathcal{Z}}~\mathcal{U}(Z|D_i) \nonumber \\ & \text{with} \ \ \ 
         \mathcal{U}(Z|D_i)=\mathbb{E}_Y[u(Y|Z,D_i)] \text{}\label{eq:8}
\end{align}
\citet{jiang2020binoculars} showed that choosing a query $\vec{z}^* \in Z^*$ is equivalent to solving $\argmax_{\vec{z}}  V(\vec{z}|D)$ where
\begin{align}
             V(\vec{z}|D_i) & =\mathbb{E}_y[u(y|\vec{z},D_i)] \nonumber \\  & + \max_{Z':|Z'|=r-1} \mathbb{E}_y[\mathcal{U}(Z'|D_i\cup{(\vec{z},y)})] \label{eq:9}
\end{align}
and that the second term of (\ref{eq:9}) is a lower bound to the second term in (\ref{eq:1}):
\begin{align}
 \max_{Z':|Z'|=r-1} \mathbb{E}_y[\mathcal{U}(Z'|D_i\cup{(\vec{z},y)})] \nonumber
 \\ \leq \mathbb{E}_{y}[\max_{\vec{z}'}\mathcal{U}_{r-1}(\vec{z}' | D_i \cup {(\vec{z},y)})] \label{eq:10}
\end{align}
Given this observation, \citet{jiang2020binoculars} proposed approximating the optimal policy (\ref{eq:1}) by optimizing its lower bound (\ref{eq:9}) which is equivalent to optimizing the batch EI known as $q$-EI. \citet{jiang2020binoculars} proposed using joint $q$-EI which is budget-unaware and scales poorly with increased dimensions \citep{wilson2018maximizing}.

%% file: proposed_approach.tex
\vspace{-1mm}
\section{PROPOSED APPROACH}\label{proposed_approach}
\begin{figure*}[h]
    \centering
    \includegraphics[width=1\textwidth]{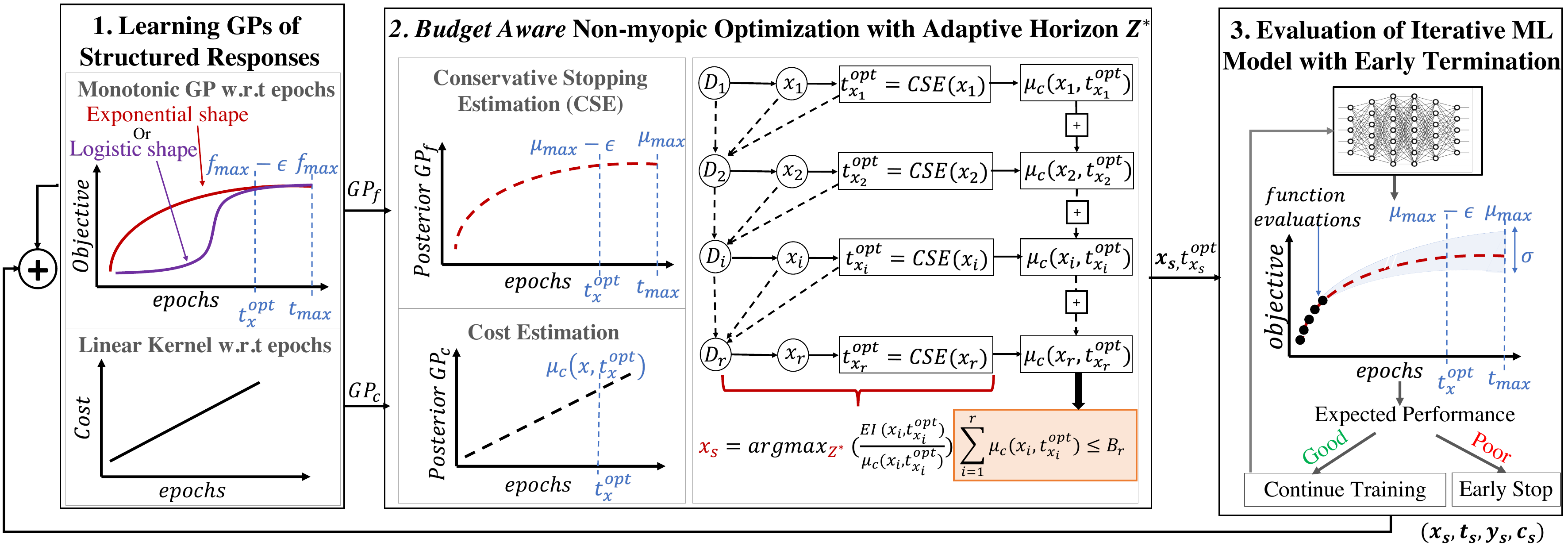}
        \vspace{-3ex}
    \caption{Overview of BAPI algorithm illustrating its three key components explained in section \ref{overview_bapi}}
    \label{fig:bapi}
\end{figure*}
In this section, we start by providing a high-level overview of the proposed BAPI algorithm and briefly explain its key components. Next, we provide complete details of each component. First, we describe how to perform an efficient budget-aware non-myopic search. Second, we explain our approach to model structured response for iterative learning which can be used to estimate conservative stopping for increased resource-efficiency.
Finally, after describing an alternative kernel for the cost model, we provide the full BAPI approach with all its component coherently put together.
\vspace{-1mm}
\subsection{Overview Of BAPI Algorithm}\label{overview_bapi}
\vspace{-1mm}
Let $GP_f$ and $GP_c$ be the surrogate probabilistic models learned given a set of observed data points $D_i$ of the objective function $f$  and the cost function $c$ respectively. Let $\mu_c$ and $\sigma_c^2$ be the predictive mean and variance of $GP_c$ and $\mu$ and $\sigma^2$ be the predictive mean and variance of $GP_f$ .\\
As shown in Figure~\ref{fig:bapi}, BAPI is a sequential
algorithm with three key components listed below:

\noindent \textbf{1. Learning Surrogate Models}\hspace{10pt} We build two surrogate models $GP_f$ for the objective function and $GP_c$ for the cost function by fitting independent GPs using queries evaluated in the past. 
We enforce shape constraint on the posterior of $GP_f$ with respect to $t$ (epoch number) to incorporate prior knowledge about the monotonicity of the function. We use a special kernel for $GP_c$ to leverage our knowldge about the variability of the cost across different HPs and its linearity with respect to $t$. 

\noindent \textbf{2. Budget Aware Non-Myopic Optimization With Adaptive Horizon}\hspace{10pt} We perform non-myopic optimization to approximate the optimal lookahead horizon $Z^*$ defined as the potential sequence of inputs that can be evaluated until their conservative stopping $t^{opt}_x$ without violating the remaining budget $B_r$:  $Z^*=\{(\vec{x}_1,t^{opt}_{\vec{x}_1}) \dots (\vec{x}_r,t^{opt}_{\vec{x}_r})\}$ such that $\sum_{\vec{z}\in Z^*} c(\vec{z})\approx\sum_{i=1}^r \mu_c(\vec{x}_i,t^{opt}_{\vec{x}_i})  \le B_r$. 
While constructing the horizon $Z^*$, each input $\vec{x}_i$ is selected based on its expected improvement. The associated conservative stopping $t^{opt}_{\vec{x}_i}$ and cost $\mu_c(\vec{x}_i,t^{opt}_{\vec{x}_i})$, are estimated upon the input selection. 

\noindent \textbf{3.  Evaluation With Early Termination}\hspace{10pt}
From $Z^*$ obtained from the second step, we select the input with highest expected improvement per unit cost \textit{at its estimated conservative stopping} for evaluation. After training the model for a fraction of the maximum number of epochs, we re-estimate the performance of the input at its conservative stopping epoch. We early terminate the training if the expected performance is poor with high certainty. 
\vspace{-1mm}
\subsection{Budget Aware BO With Adaptive Horizon}\label{nonmyopic-opt}
\vspace{-1mm}
Approximating the non-myopic optimization with the batch expected improvement $q$-EI, where the batch size $q$ is equal to the horizon of the lookahead optimization $r$, is an efficient approach \citep{jiang2020binoculars}. However, the \textit{joint} $q$-EI via reparametrization trick and Monte Carlo sampling proposed in \citet{wang2016parallel} and used in \citet{jiang2020binoculars} requires the size of the batch to be fixed and solves a \textit{joint one-shot} optimization problem of $(d\times q)$ dimensions.

\textbf{Challenges}\hspace{10pt} 1) In the context of budgeted non-myopic optimization, the horizon of remaining queries $r$ is unknown: it depends on remaining budget $B_r$ and expected costs of horizon queries $\vec{z}_i, i\in \{1 \dots r\}$. An efficient method should allow the horizon to be adaptive to the budget. Therefore, the \textit{joint} $q$-EI is not a suitable solution. 2) Given an optimization problem with reasonable medium-size dimension and a medium length horizon, the dimensionality of the joint optimization problem can significantly increase. \citet{wilson2018maximizing} showed that the performance of the joint $q$-EI deteriorates for large optimization dimension.

\noindent\textbf{Proposed Alternative}\hspace{10pt} To overcome the above two challenges, we propose to employ the sequential greedy $q$-EI via reparametrization trick and Monte Carlo sampling proposed in \citet{wilson2018maximizing}. 
\citet{wilson2018maximizing} showed that $q$-EI is a submodular acquisition function, which guarantees a near-optimal maximization via a sequential greedy approach. This incremental version of the acquisition function has several distinct advantages over the joint one: 1) It is amenable to an adaptive horizon, where we can stop adding points to the batch based on the remaining budget. 2) It is more efficient and produces better performance when the value of $d\times q$ is high \citep{wilson2018maximizing}.
After the batch approximation returns a sequence (horizon) of inputs, we select one input to query its expensive function evaluation. We discuss an input selection strategy, that is relevant to iterative machine learning models optimization, in section \ref{proposed_algo}. Note that our approximation can clearly extend to the use of any other batch acquisition function that satisfies the submodularity condition and have a sequential greedy approach, namely the $q$-UCB and $q$-PI \citep{wilson2018maximizing}. 

It is important to highlight that in practice, our approach can naturally extend to parallel BO evaluation (batch setting). The user can choose more than one point from the approximated horizon and evaluate them in parallel as long as the horizon length is fairly larger than the number of selected points for evaluation. Even though in this paper we focus on the sequential setting, we enable this option in our implementation. We provide a summary of the budget-aware BO approach in Algorithm \ref{alg:pseudocode}.
\begin{algorithm}[h!]
\footnotesize
\caption{\small{Budget Aware Non-myopic BO}}
\label{alg:pseudocode}
 \textbf{Input:}  $\mathcal{Z}$, $f(\vec{z}), c(\vec{z})$,  models $GP_f,GP_c $, utility function $u(y| \vec{z}, D)$, a total budget $B_T$ \\
\textbf{Output:} $D, \vec{z}^*, f(\vec{z}^*)$
\begin{algorithmic}[1]
    \STATE Initialize the remaining budget $B_r \leftarrow B_T$
    \WHILE{$B_r \geq 0$:}
    \STATE Approximate the optimal horizon via adaptive optimal batch computation $Z^*$ of size $r$ such that $\sum_{i=0}^{r} \mu_c(\vec{z}_i) \leq B_r$
    \STATE Select a candidate input $\vec{z}^*\in Z^*$\ and observe its evaluation $f(\vec{z}^*)=y^*$ and cost $c(\vec{z}^*)=y_c^*$
    \STATE Update the remaining budget $B_r \leftarrow B_r -c(\vec{z}^*)$
    \STATE Update data $D=D\cup\{(\vec{x^*},y^*,y_c^*)\}$
    \ENDWHILE
\end{algorithmic}
\end{algorithm}
\vspace{-1.5mm}
\subsection{Structured Responses}
In this section, we describe our proposed approaches to leverage prior knowledge about the structure of the responses, namely, the monotonicity and shape of the function $f$ and the linearity of the cost $c$.

We propose to use a GP with monotonicity constraint over the $t$ variable to model the function $f$. Recent work \citep{agrell2019gaussian} proposed an efficient approach to introduce linear operator inequality constraints to GPs. 
Let $f$ be the function modeled by a GP  and $\mathcal{L}$ be a linear operator. The proposed approach enables the posterior prediction to account for inequality constraints defined as
$
     a(\vec{z}) \leq \mathcal{L}f(\vec{z}) \leq b(\vec{z}).
$
The derivative operator is a linear operator. Hence, to apply monotonicity, this condition can be seen as the partial derivative of the model of $f$ with respect to $t$ is positive.   
For this condition to hold, \citet{agrell2019gaussian} proposed to define a set of \textit{virtual observation locations} $Z^v=\{\vec{z_1}^{v},\dots,\vec{z_s}^{v}\}$ where the condition is guaranteed to be satisfied. 

The posterior predictive distribution of the monotonic GP is $\vec{f}^{*} |Y, C$, 
which is the distribution of $\vec{f}^{*} = f(\vec{z}_{*})$ for some new inputs $\vec{z}_{*}=[\vec{x}_*,t_*]$, conditioned on the observed data $Y$ and the constraint $C$ defined as $a(Z^{v}) \leq \mathcal{L}f(Z^{v}) \leq b(Z^{v})$.
The final derivation of the predictive distribution, provided by \citet{agrell2019gaussian}, is defined as follows: 
 \begin{align*}
       & \textbf{\textup{f}}^{*} | Y, C \sim \mathcal{N}(\mu^{*} + A(\textbf{C} - \mathcal{L} \mu^{v}) + B(Y - \mu), \Sigma ) \\
       & \textbf{C} = \widetilde{C} | Y, C \sim \mathcal{TN}( \mathcal{L} \mu^{v} + A_{1}(Y - \mu), B_{1}, a(Z^{v}), b(Z^{v}))
    \end{align*}
    where $\mathcal{TN}(\cdot, \cdot, a, b)$ is the truncated Gaussian $\mathcal{N}(\cdot, \cdot)$ conditioned on the hyper-rectangle 
    $[a_{1}, b_{1}] \times \cdots \times [a_{k}, b_{k}]$, $\mu^v=m(Z^v), \mu^*=m(\vec{z}_*), \mu=m(Z)$. The definition of the matrices $A,B,A_1,B_1$ and $\Sigma$ can be found in Appendix \ref{appendixA}.
 The computation of the posterior of the monotonic GP requires the definition of derivatives of the kernel function. In this work we consider monotonicity with respect to one dimension $t$. Therefore, we need the first order derivatives.

In cases where the function is known to be exponentially decaying (e.g., neural network training), the kernel over dimension $t$ should be defined as an ED kernel. However, in cases where the shape of learning curve is monotonic but not necessarily  exponentially decaying (e.g., cumulative and average reward for RL models), an RBF kernel with monotonicity over dimension $t$ should be used. Leveraging monotonicity in the modeling allows flexibility and the generalization of our approach for several types of ILs. We provide the derivatives for both kernels and the details about the specification of the location of virtual observations with each kernel in Appendix \ref{appendixA} and in our implementation. We additionally provide insights about the efficient posterior computation of the monotonic GP. For more details, we refer the reader to \citet{agrell2019gaussian}.

\noindent\textbf{Conservative Stopping Estimation}
Previously proposed BO approaches for HPO consider a maximum number of epoch $t_{max}$ at which the objective function will reach its best value. However, in practice, different HPs do not need to necessarily run to the maximum number of epochs to reach their optimal value as the objective stops improving (reaches a plateau) \citet{kaplan2020scaling}. Therefore, running them for longer epochs can be a waste of limited resource budget with  diminishing returns.
Existing work proposed early stopping of HPs based on their performance compared to previously evaluated data points \citep{li2017hyperband,dai2019bayesian, swersky2014freeze} or based on the expected improvement per unit cost \citep{nguyen2020bayesian} which leads to the selection of very low number of epoch due to the high cost of $t_{max}$.
We propose to define a conservative stopping $t_{\vec{x}}^{opt}$ for each HP $\vec{x}$ as the smallest number of epoch needed to reach the best function value at $\vec{x}$. Our approach enables the estimation of when a learner becomes $\epsilon$-close to its asymptotic value.
To the best of our knowledge, no previous work used the estimation of the function values at another location to reason about the HP selection and optimal early termination before reaching that epoch.
The problem of estimating $t_{\vec{x}}^{opt}$ for a HP $\vec{x}$ based on the GP posterior is defined as below and efficiently solved using binary search.
\begin{align}
  & t_{\vec{x}}^{opt} \leftarrow \arg min_{t \in [t_{min},t_{max}]} ~ t \\  & s.t  ~ \mu(\vec{x},t_{max}) - \mu(\vec{x},t)\leq \epsilon \nonumber
\end{align}
\textbf{Cost Modeling}\hspace{10pt}
The cost prediction is an important component in our algorithm. Therefore, it is important to have an accurate and informative model for the cost. We propose to model the cost by an independent Gaussian process $GP_c$ that captures two important characteristics: 1) The cost of the training of different HPs for the same number of iterations $t$ can vary significantly. 2) The cost of training of a fixed HP $\vec{x}$ increases linearly with the number epochs $t$.
We propose to use the product kernel 
$K_c([\vec{x},t],[\vec{x'},t'])=K_{c_x}(\vec{x,x'})\times K_{c_t}(t,t')$
, where $K_{c_x}$ is an RBF kernel over $\vec{x}$ and $K_{c_t}$ is a linear kernel over $t$. 
Note that previous work assumes the cost is same for different HP $\vec{x}$ and linear with respect to $t$. This might lead to an inaccurate estimation of the cost especially if some of the dimensions of $\vec{x}$ represent architectural variables (e.g number of layers, number of hidden nodes etc.) 
\vspace{-1mm}
\subsection{Budget-Aware Planning For Iterative Learners (BAPI)}\label{proposed_algo}
In this section, we describe the overall budget-aware non-myopic BO algorithm for HPO of  iterative learners.  The main idea is to use the reparametrized iterative greedy q-EI proposed in \citet{wilson2018maximizing} to approximate the optimal sequence of selections with respect to the available budget. q-EI will have an adaptive batch size with budget exhaustion as a stopping criteria. 
We propose to adaptively add inputs to the horizon based on their expected improvement at their conservative stopping iteration without normalizing the utility function by the cost during the optimization. The details of execution can be found in the Non-Myopic Optimization (NMO) function described in Algorithm \ref{planning}. This function returns a set of inputs representing the optimal horizon $Z^*$. 

\textbf{Input Selection From Horizon}\hspace{10pt}
Given the set of inputs $Z^*$, how to select the next input to evaluate? We propose to select the input with the highest immediate expected reward per unit cost \textit{at its conservative stopping iteration}. We note here that this is \textit{different} from optimizing the utility function per unit cost and the issue of selecting low non-informative number of iterations would not arise. In this case, the number of iterations is already fixed to an optimal high value for each input $\vec{x^*}$.

\textbf{Early Termination}\hspace{10pt}
After selecting the next candidate HP to evaluate, the function evaluation will return a $y_t$ value after each epoch. Based on the function values of the initial $p$ epochs, we can re-estimate the final performance of $\vec{x}$ and its new conservative stopping $t_{\vec{x}}^{opt_n}$. The algorithm makes a decision to continue model training with the current HP or early-stop it.
If both 1. $\mu(\vec{x},t_{\vec{x}}^{opt_n}) \leq y_{{best}}$,
and
2. $\sigma(\vec{x},t_{\vec{x}}^{opt_n}) \leq \tau \sigma(\vec{x},t)$, then
model training will be early stopped in epoch t, otherwise, the conditions will be verified again after running another set of $p$ epochs or when it reaches the estimated $t_{\vec{x}}^{opt_n}$, whichever happens earlier. 
The first condition will recommend stopping the training if the predicted function value at $t_{\vec{x}}^{opt_n}$ will not be higher than the current best value achieved across all evaluated HP. The second condition recommends the early stopping only if the uncertainty of the model about the predicted function value at the estimated conservative stopping is no more than a factor $\tau\geq 1$ of the uncertainty of the model about the last evaluated epoch. 
In another word, condition 2 will prevent the early stopping if the model is not certain enough about its prediction of the function value at $t_{\vec{x}}^{opt_n}$. Algorithm \ref{evaluate} summarizes evaluation with early termination.

\begin{algorithm}[H]
\footnotesize
\caption{\small{BAPI}}
\label{alg:full}
\textbf{Input}: $\mathcal{X}$; $f(\vec{x},t)$;$c(\vec{x},t)$; $t_{max}$; $B_{T} $\\
\textbf{Output:}$\vec{x}^*,t_{\vec{x}^*}^{opt},f(\vec{x}^*,t_{\vec{x}^*}^{opt})$
\begin{algorithmic}[1] 
\STATE Initialize with $N_0$ initial points 
\STATE Fit the models: $GP_f,GP_c$ 
\STATE $B_{r} \leftarrow B_{T}- \sum_{i=0}^{{N_0}} c(\vec{x}_i,t_{\vec{x}_i})$ 
\WHILE{$B_{r}>0$} 
\STATE \hspace{-2mm}\textit{\# Find the budget constrained horizon and their corresponding conservative stopping}\\
\hspace{-2mm}$Z^*:\hspace{-1mm}\{(\vec{x}_{1},t_{\vec{x}_{1}}^{opt})\cdots(\vec{x}_{r},t_{\vec{x}_{r}}^{opt})\} \leftarrow NMO(GP_f,GP_c,B_{r})$
\STATE \hspace{-2mm}\textit{\# Select one point for evaluation}\\
\hspace{-2mm} $ \vec{x},t_{\vec{x}}^{opt} \leftarrow \arg max_{Z^*} ~ \frac{EI(\vec{x}_i,t_{\vec{x}_i}^{opt})}{\mu_c(\vec{x}_i,t_{\vec{x}_i}^{opt})} $ \\
\STATE \hspace{-2mm}$ \vec{y},\vec{y}_c \leftarrow Evaluate( f(\vec{x},t_{\vec{x}}^{opt}))$
\STATE \hspace{-2mm}Aggregate data: $\mathcal{D} \leftarrow \mathcal{D} \cup \{(\vec{x}, \vec{y},\vec{y}_c )\}$ 
\STATE \hspace{-2mm}Update Models: $GP_f,GP_c$ 
\STATE  \hspace{-2mm}$B_{r}\leftarrow B_{r}- c(\vec{x},t_{\vec{x}})$
\ENDWHILE
\end{algorithmic}
\end{algorithm}
\vspace{-6mm}
\begin{algorithm}[H]
\footnotesize
\caption{\small{Conservative Stopping Estimation}}
\label{optS}
$ConservativeStopping(GP_f,\vec{x})$
\begin{algorithmic}[1] 
\STATE $t_{\vec{x}}^{opt} \leftarrow \arg min_{t \in [t_{min},t_{max}]} ~ t$ 
\STATE  \quad s.t $\mu(\vec{x},t_{max})-\mu(\vec{x},t)< \epsilon$
\STATE \textbf{Return} $t_{\vec{x}}^{opt}$
\end{algorithmic}
\end{algorithm}
\vspace{-6mm}
\begin{algorithm}[H]
\footnotesize
\caption{\small{Adaptive Horizon $q$-EI AFO}}
\label{planning}
$NMO(GP_f,GP_c,B_{r})$
\begin{algorithmic}[1] 
\STATE $Z^*=\{\}$
\WHILE{ $B_{r}>0$}
\STATE \textit{\# Add $\vec{x}$ based on the highest number of epochs}\\
$ \vec{x} \leftarrow \arg max_{\vec{x}\in \mathcal{X}}q$-EI$(\vec{x},t_{max}) $ 
\STATE \textit{\# Estimate the conservative stopping for $\vec{x}$}\\
$t_{\vec{x}}^{opt} \leftarrow ConservativeStopping(GP_f,\vec{x})$ 
\STATE \textit{Deduct estimate cost at $t_{\vec{x}}^{opt}$ from budget}\\
$B_{r}\leftarrow B_{r }- \mu_c(\vec{x},t_{\vec{x}}^{opt})$
\STATE $Z^*=Z^* \cup \{(\vec{x},t_{\vec{x}}^{opt}\}$
\ENDWHILE
\STATE \textbf{Return} $Z^*$
\end{algorithmic}
\end{algorithm}
\vspace{-6mm}
\begin{algorithm}[H]
\footnotesize
\caption{\small{Evaluate Function}}\label{evaluate}
$Evaluate(f(\vec{x},t_{\vec{x}}^{opt}))$
\begin{algorithmic}[1] 
\STATE $t\leftarrow p$
\WHILE{$t\leq t_{max}$ and $Continue$}
\STATE $\vec{y}=f(\vec{x},t)$ ; $\vec{y}_c=c(\vec{x},t)$
\STATE $t_{\vec{x}}^{opt_n} = ConservativeStopping(GP_f,\vec{x})$
\STATE if $\mu(\vec{x},t_{\vec{x}}^{opt_n}) \leq y_{{best}}$ and $\sigma(\vec{x},t_{\vec{x}}^{opt_n}) \leq \tau \sigma(\vec{x},t) $:\\
\quad $Continue \leftarrow False$
\STATE else: \\
 \quad $Continue \leftarrow True$ \\
 \quad $t \leftarrow min(t_{\vec{x}}^{opt_n},t+p) $
\ENDWHILE
\STATE  \textbf{Return}  $\vec{y},\vec{y}_c$
\end{algorithmic}
\end{algorithm}

\textbf{Data Points Selection From Learning Curve}\hspace{10pt}
Iterative machine learning models evaluated with an input configuration $\vec{x}^*$ and a number of epochs $t^*$ return a vector of $t^*$ function values and a vector of $t^*$ cost values associated with each iteration $t\leq t^*$. Most of existing work, do not utilize these data points and use only the function value at the last epoch. However, leveraging part of this data can help the learning of the monotonic shape of the objective function and result in a more accurate extrapolation. We select, from each curve, the points with the highest model uncertainty(variance) following the approach proposed in \citet{nguyen2020bayesian}.

\textbf{Practical Considerations}\hspace{10pt}
Considering a perfect model of the function, querying a complete lookahead horizon in each iteration would be optimal. However, as pointed out by previous work on non-myopic BO \citep{jiang2020binoculars,lee2021nonmyopic,lee2020efficient,yue2020non}, the model is usually uncertain about long term predictions. Consequently querying a long horizon can hurt the optimization by evaluating misleading points and causing a higher computational cost. Therefore, we follow previous work \citep{astudillo2021multi} and set a maximum horizon length as an additional stopping condition to the size of the horizon. Given this mitigation, we expect that the horizon adaptation to the budget to occurs depending on the remaining budget. Additionally, selected points would always be within the limits of the remaining budget.

\textbf{Cost of a Restarted Hyperparameter}
In iterative learning, the optimization algorithm might select a configuration that was previously evaluated for a lower number of epochs. However, the cost will always be estimated with respect to the evaluation iteration. For an accurate optimization, our algorithm handles this special case by assigning a cost that only reflect the additional epochs to be run. This is accounted for in the non-myopic optimization function, input selection, and budget deduction after function evaluation. 

%% file: Related_work.tex
\section{RELATED WORK}
Our problem setting and the proposed BAPI solution have many intersections with previous work in hyperparameter optimization,
Bayesian optimization, non-myopic optimization, and sequential decision making which we attempt to summarize here.

\textbf{HPO/BO For Iterative Learning}\hspace{10pt}
\cite{domhan2015speeding} proposed learning curve prediction in order to allow early termination of non-promising candidates.
This approach utilizes approximate Bayesian inference w.r.t.\ a pre-defined finite set of learning curve models to perform extrapolation to a fixed horizon.
\cite{klein2016learning} built on this method by showing Bayesian neural networks could be used for learning curve prediction.
\cite{swersky2014freeze} proposed a hierarchical GP model for HP tuning that includes learning curve prediction upon which decisions for exploration (freeze current and test new candidate) and exploitation (thaw current and continue learning) are based.
More recently, \cite{dai2019bayesian} proposed an optimal stopping procedure for increasing the sample efficiency of BO and showed competitive performance with \cite{domhan2015speeding} for iterative learners. While this procedure obtains theoretical guarantees, 
it must generate a sample of large size from the GP in order to make a reliable decision.
Moreover, solving for the stopping rule requires an approximate backward induction technique after each epoch.
Our proposal for early termination is conceptually simpler and far less computationally demanding.

The method of \cite{lu2019hyper} considers a finite set of learners modeled with freeze/thaw \citep{swersky2014freeze} selecting between exploration and exploitation with a heuristic 
$\epsilon$-greedy rule. 
\cite{wu2020practical} extend the knowledge gradient acquisition function \citep{frazier2008knowledge} to trace-valued observations that occur in multi-fidelity applications. 
BOIL \citep{nguyen2020bayesian} also considers trace-valued observations, but compresses the trace via learned weighted sum as well as adding carefully-chosen intermediate trace values as observations.
The setting for BOIL includes reinforcement learning problems with reward functions taking non-exponentially decaying shapes. 
Our BAPI approach uses a product kernel to jointly model correlations between HPs and epochs within the iterative training procedure.

\citet{kandasamy2017multi} developed BOCA, an extension of UCB to general multi-fidelity BO setting.

We note that there exist orthogonal approaches that focus on the ability to extrapolate responses based on smaller datasets where the cost is varied based on the size or fraction of the dataset used for training. These methods propose algorithms based on multi-task BO \citep{swersky2013multi,klein2017fast} and importance sampling \citep{ariafar2021faster}.

\textbf{Non-Myopic Policies} While there were early attempts at non-myopic selection for length-two horizons \citep[e.g.,][]{osborne2010bayesian}, most work on proposing practical methods for longer horizons is very recent.
\citet{wu2019practical} developed gradient estimates for two-step EI admitting gradient-based search for the optimal two-step
selection.
\citet{lam2016bayesian} utilized a Markov decision process (MDP) formalism and performed rollouts with a predefined base policy to estimate the value function.
GLASSES \citep{gonzalez2016glasses} approximates the solution for the optimal non-myopic selection by a combination of approximate integration given future selections and approximating the future selections by a diversity-promoting batch selection procedure from \citet{gonzalez2016batch}).

Closely-related to our work is BINOCULARS \citet{jiang2020binoculars} which was discussed in Section \ref{background}.
The major differences to our proposed BAPI approach are that
(a) BINOCULARS uses \textit{joint} batch expected improvement $q$-EI \citet{wang2016parallel} while we use the sequential greedy selection \citet{wilson2018maximizing},
(b) BINOCULARS uses a fixed horizon that is budget agnostic while we use a budget-adaptive horizon,
and
(c) BINOCULARS does not take cost into account when returning its non-myopic selected query while our method does factor cost in.
\citet{lee2021nonmyopic} consider the general cost-aware setting and frame the problem as a constrained MDP.
Their method approximately solves the intractable problem by performing rollouts of a base policy that does not adapt the horizon to the budget.
Moreover, it's base policy is normalized by cost leading to a bias towards low-cost queries. 
Building upon the efficient one-shot multi-step tree approach of
\citet{jiang2020efficient},
\citet{astudillo2021multi} introduce cost-modeling and develop a budget-aware method.
However, this method's adaptation to the horizon is post-hoc in the sense that the horizon has to be fixed in advance and cannot be adaptive to the remaining budget due to utility function formulation and optimization. This leads to an unnecessary higher dimensional optimization and a manual zero padding technique to handle cases where the selected horizon violates the remaining budget.

\textbf{Bandit Algorithms}
Given that the objective in (\ref{eq:problem}) is equivalent to optimizing for simple regret,
there is a large amount of relevant work within the multi-armed bandits literature.
\citet{audibert2010best} developed the upper confidence bound exploration (UCB-E) policy, for the best arm identification (BAI) in the budgeted setting by providing conditions under which simple regret decays exponentially with increasing budget.
\citet{hoffman2014correlation} considered linear bandits and proposed 
the BayesGap algorithm which is an exploration policy within budget constraints. 
Later, \citet{jamieson2016non} analyzed successive halving as an instance of non-stochastic multi-armed bandits in the setting where the budget is greater than the number of learners.
HyperBand \citep{li2017hyperband} is an implementation of successive halving running this algorithm in multiple successive rounds 
and is a very general algorithm for HPO including non-iterative learners. 
Most recently, BOHB \citep{falkner2018bohb} modified HyperBand by utilizing BO within the successive halving procedure which guides the selection process for the learners that will be trained for longer budgets.

%% file: experiments.tex
\section{EXPERIMENTS AND RESULTS}\label{experiments_results}
In this section, we first provide details about our experimental setup. Next, we evaluate the performance of BAPI approach and compare it to several state-of-the-art baselines.
\vspace{-0.5mm}
\begin{figure*}[h]
    \centering
    \begin{minipage}{0.245\textwidth}
        \centering
        \includegraphics[width=1\textwidth]{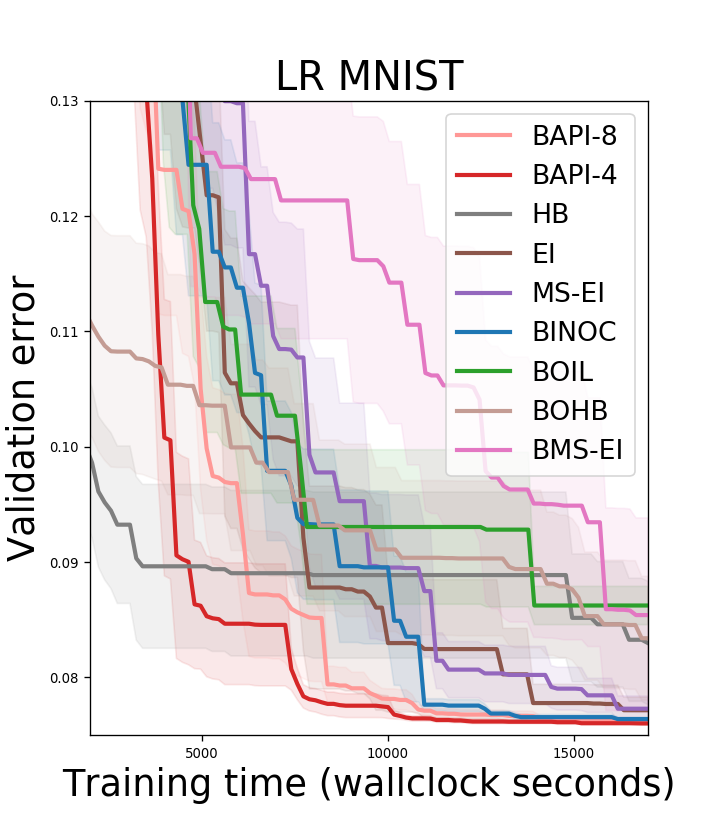} 
    \end{minipage}\hfill
    \begin{minipage}{0.245\textwidth}
        \centering
        \includegraphics[width=1\textwidth]{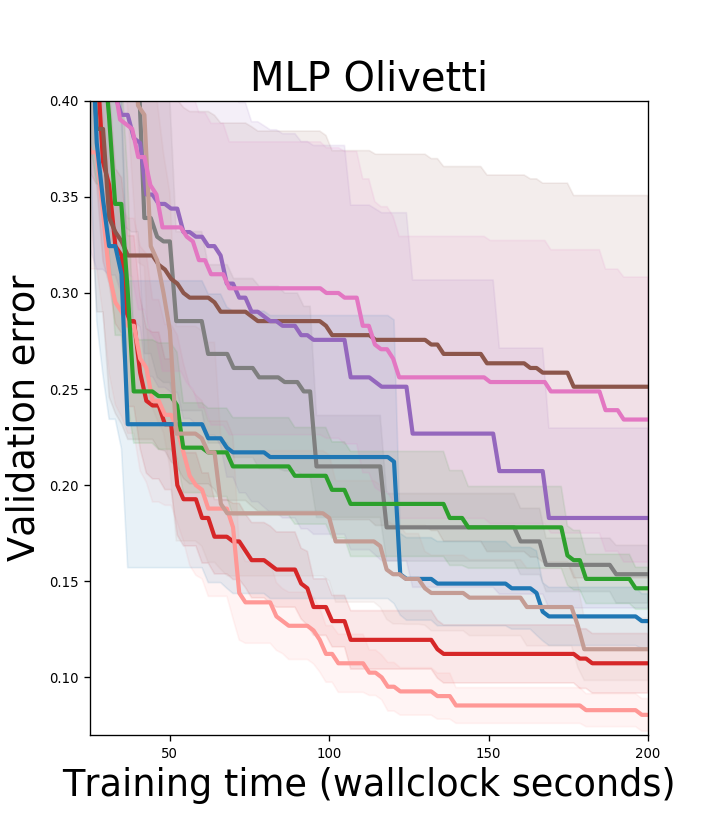} 
    \end{minipage} \hfill
    \begin{minipage}{0.245\textwidth}
        \centering
        \includegraphics[width=1\textwidth]{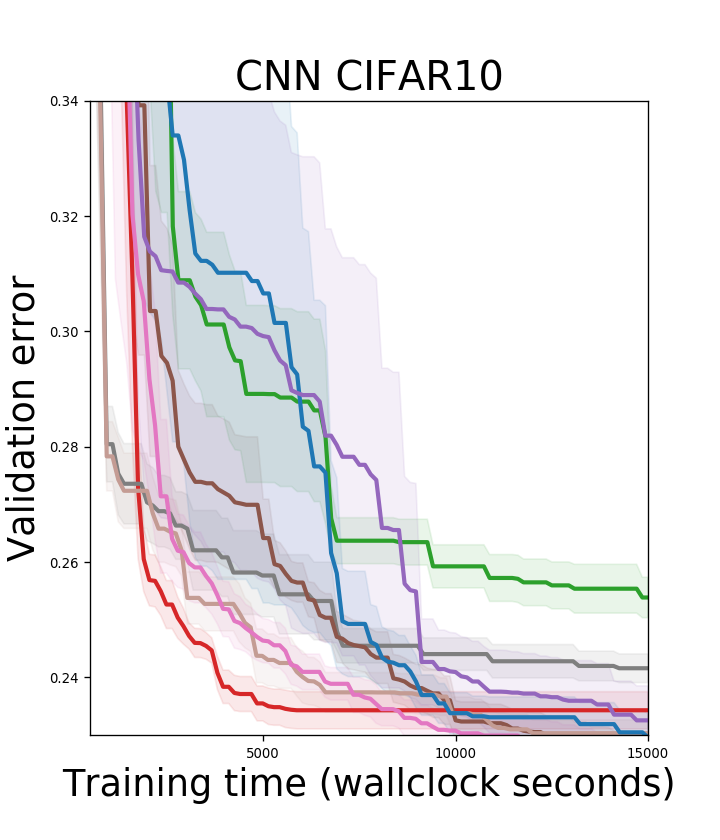} 
    \end{minipage}\hfill
    \begin{minipage}{0.245\textwidth}
        \centering
        \includegraphics[width=1\textwidth]{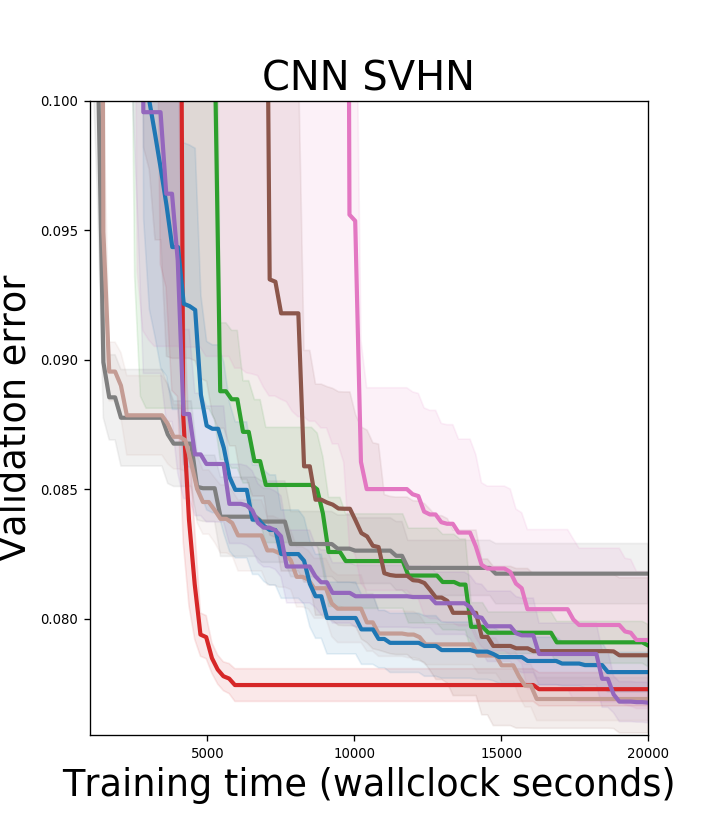} 
    \end{minipage} 
\caption{Results of validation error $\pm$ standard error for different baselines and our proposed approach on multiple iterative learners against training budget.}
\label{experiments_figure}
\end{figure*}
\begin{table*}[h]
\centering
\caption{Average ranking of BAPI and baseline methods across all experiments.}
\label{tablerank}
\begin{tabular}{lllllllll}  
\toprule
 Algorithm & BAPI & HB & BOHB & EI & MS-EI & BINOC  & BOIL & BMS-EI\\
\midrule
Average ranking & { 2.9 $\pm$0.51} &  {6.1$\pm$0.51} & 3.6$\pm$0.67 & 4.9$\pm$0.53 &  3.4$\pm$0.50 &  4.3$\pm$0.38 &  6.3$\pm$0.63 &  4.2$\pm$0.73 \\
\bottomrule
\end{tabular}
\label{tablerun}
\end{table*}
\noindent \textbf{Baselines.} 
We evaluate state-of-the-art baselines, described in the related work: from cost-aware non-myopic BO literature \texttt{BMS-EI}\footnote{github.com/RaulAstudillo06/BudgetedBO} \citep{astudillo2021multi},  from non-myopic BO BINOCULARS \texttt{(BINOC)} \citep{jiang2020binoculars}\footnote{\label{enbo_code}github.com/shalijiang/bo/tree/main/enbo}and \texttt{MS-EI} \citep{jiang2020efficient}\footnoteref{enbo_code}, from general BO literature \texttt{EI} \citep{jones1998efficient}, from HPO for iterative learners literature \texttt{BOHB} \citep{falkner2018bohb}\footnote{\label{HpBandSter}github.com/automl/HpBandSter} and HyperBand \texttt{(HB)} \citep{li2017hyperband}\footnoteref{HpBandSter}, from multi-fidelity BO for HPO literature \texttt{BOIL} \cite{nguyen2020bayesian}\footnote{github.com/ntienvu/BOIL}.
Each baseline implementation uses settings recommended by the original authors and publicly available code. 
We also evaluated  GLASSES \citep{gonzalez2016glasses}\footnoteref{enbo_code} and random search. However, both of them performed always poorly when compared to all other baselines. Therefore, for clarity of the figures, we do no report them. We note that previous work on
non-myopic BO do not include HB and BOHB as baselines but given their competitive performance in iterative learning settings, we recommend they become standard in future work in this problem setting.

\textbf{Experimental Setup}\hspace{10pt} All experiments were averaged over 10 runs with different random seeds. The code of our BAPI implementation is publicly available\footnote{\label{bapi_code}github.com/belakaria/BAPI}.
We considered several  state-of-the-art HPO benchmarks: 1) Logistic regression with MNIST dataset; 2) Multi-layer perceptron with Olivetti dataset; 3) Multi-layer perceptron with Covtype dataset ; 4) Fully connected network with MNIST dataset with two different $t_{max}$ setups  5) CNN on image dataset CIFAR10 with two different $t_{max}$ setups; 6) CNN on SVHN dataset with two different $t_{max}$ setups; 7) Resnet on CIFAR100 dataset; 8) A Dueling DQN (DDQN) agent in the CartPole-v0 environment; 9) An Advantage Actor Critic (A2C) agent in the Reacher-v2 environment; and 10) An Advantage Actor Critic (A2C) agent in the InvertedPendulum-v2 environment.
Full experimental details are listed in Appendix \ref{experiments_details}. We report the validation error as the evaluation metric for consistency across datasets.
We evaluate two different variants of our algorithm: BAPI-4 and BAPI-8, where the maximum horizon is set to 4 and 8 respectively. BAPI-8 was evaluated on two benchmarks (LR with MNIST and MLP with Olivetti) to demonstrate the effect of varying the maximum horizon on the performance.
The uncertainty threshold $\tau$ is set to 2 for all experiments.
The parameter $p$ is set to 20\% of the maximum number of epochs for all experiments except for CNN-SVHN, where it is set to 10\% due to the high cost of each epoch. We select at most three data points from each learning curve.\\
\textbf{Setting $\epsilon$}\hspace{10pt} For experiments 1 to 7,
$\epsilon$ is set to 0.01 (interpreted as at most 1\% degradation in accuracy) except for CNN-SVHN, where it is set to 0.005 due the small variation in the validation error. In the case of a loss/reward function where $\epsilon$ cannot be easily set (e.g experiments 8 to 10), it is automatically set as the smallest degradation in the function value at $t_{max}$ in the evaluated data: $\epsilon = min \{f(x,t_{max})-f(x,t) ~ \forall (x,t) \in D\} $. In general, $\epsilon$ is not required to be fixed. A strategy for updating it, that balances exploration and exploitation (e.g., a wider value and therefore earlier stopping in the beginning), can be set by the practitioner and interfaced with our code easily.
 {\em We provide additional results and discussion in {\bf Appendix \ref{experiments_details}}}.

\noindent\textbf{Results and Discussion}\hspace{10pt} Figure \ref{experiments_figure} shows the results (best validation error as a function of wall-clock time) of all methods on four HPO tasks. We make the following observations.
{\bf 1)} BAPI identifies better candidates with less total cost than the baselines
due to its ability to plan selections while accounting for budget 
and early terminating non-promising candidates.
{\bf 2)}
A longer horizon for BAPI was tested on the relatively cheaper experiments LR MNIST and MLP Olivetti datasets and shows some
performance degradation (LR MNIST) and some improvement (MLP Olivetti) suggesting that optimal horizon length is
problem-dependent but clearly helpful in some cases. {\bf 3)} BMS-EI had an unstable performance and was not able to uncover good candidate in several experiments. We speculate that it is due to the approach being conservative about which points would satisfy the remaining horizon constraint.
{\bf 4)}
HB and BOHB can identify good candidates faster than most algorithms in the beginning, mainly because their strategy forces initial evaluations to be low-epoch trained runs. 
In the mid-range, their performance slows down, perhaps
a consequence of their exploitation behavior and reliability on successive halving which might limit their extrapolation ability and stop promising candidates very early.
Similar analysis has been reported in previous work \citep{dai2019bayesian}.
With longer search times, BOHB can catch up. However, both BOHB and HB performance degrades significantly in RL settings since successive halving cannot extrapolate accurately when the function might take a sigmoid or logit shape.
{\bf 5)} BINOCULARS and MS-EI, are slower to uncover promising candidates due to spending more budget in evaluating all selected candidates to the maximum number of epochs.
However, they both arrive at a competitive performance towards the end that can be attributed to their planning capabilities.
{\bf 6)} BOIL is worse than most baselines across all benchmarks. As discussed in Section \ref{intro}, BOIL selects the next candidates by weighting EI by the cost and might suffer from the cost miscalibration pathology. Similar observations were made in \citet{astudillo2021multi}.

Additional results included in Appendix \ref{experiments_details} show that BAPI becomes less competitive
when the total budget is significantly increased, but also provides results suggesting that performance loss can be brought back
by adjusting the pruning behavior, a topic of on-going work. 

We compared our approach to a wide range of baselines on 13 different experiments. All the baselines had inconsistent performance across the different experiments while our algorithm performed fairly well across all of them. The gain was significant in the case of limited budget, which is the desired behavior since a planning approach is more needed when the budget is limited. The gain was less significant in experiments with higher budgets but our approach was still competitive. Therefore, in Table \ref{tablerank}, we provide the average ranking of each algorithm over all experiments based on their final performance.  

%% file: appendix.tex
\section{Details of the monotonic Gaussian process }\label{appendixA}

The posterior predictive distribution of the monotonic GP is $\vec{f}^{*} |Y, C$
which is the distribution of $\vec{f}^{*} = f(\vec{z}_{*})$ for some new inputs $\vec{z}_{*}=[\vec{x}_*,t_*]$, conditioned on the observed data $Y$ and the constraint $C$ defined as $a(Z^{v}) \leq \mathcal{L}f(Z^{v}) \leq b(Z^{v})$.
The final derivation of the predictive distribution is defined as follow: 
 \begin{equation}
        \textbf{\textup{f}}^{*} | Y, C \sim \mathcal{N}(\mu^{*} + A(\textbf{C} - \mathcal{L} \mu^{v}) + B(Y - \mu), \Sigma )
    \end{equation}
    \begin{equation}
        \label{eq:posterior_2}
        \textbf{C} = \widetilde{C} | Y, C \sim \mathcal{TN}( \mathcal{L} \mu^{v} + A_{1}(Y - \mu), B_{1}, a(Z^{v}), b(Z^{v}) )
    \end{equation}
    where $\mathcal{TN}(\cdot, \cdot, a, b)$ is the truncated Gaussian $\mathcal{N}(\cdot, \cdot)$ conditioned on the hyper-rectangle 
    $[a_{1}, b_{1}] \times \cdots \times [a_{k}, b_{k}]$, $\mu^v=m(Z^v), \mu^*=m(\vec{z}_*), \mu=m(Z)$. The matricies $A,B,A_1,B_1$ and $\Sigma$ are defined as follow:
        \begin{align}
            & A_{1} = (\mathcal{L} K_{Z^{v}, X})(K_{Z, Z} + \sigma^{2}I)^{-1}\\
            & A_{2} =  K_{\vec{z}_*, Z}(K_{Z, Z} + \sigma^{2}I)^{-1} \\
            & B_{1} = \mathcal{L} K_{Z^{v}, Z^{v}} \mathcal{L}^{T} + \sigma^{2}_{v} I - A_{1} K_{Z, Z^{v}}\mathcal{L}^{T} \\
            & B_{2} = K_{\vec{z}_*, \vec{z}_*} - A_{2} K_{Z, \vec{z}_*} \\ 
            & B_{3} = K_{\vec{z}_*, Z^{v}}\mathcal{L}^{T} - A_{2} K_{Z, Z^{v}} \mathcal{L}^{T} \\ 
            & A = B_{3}B_{1}^{-1} \\
            & B = A_{2} - AA_{1} \\
            & \Sigma = B_{2} - AB_{3}^{T}
        \end{align}
        
    Additionally, the probability that the unconstrained version of $\textbf{C}$ falls within the constraint region, $p(C | Y)$, is defined as follow:
    \begin{equation}
        \label{eq:constrprob}
        p(C | Y) = p \left( a(Z^{v}) \leq \mathcal{N}( \mathcal{L} \mu^{v} + A_{1}(Y - \mu), B_{1})  \leq b(Z^{v}) \right) 
    \end{equation}
    and the unconstrained predictive distribution is 
    \begin{equation*}
        \textbf{\textup{f}}^{*} | Y \sim \mathcal{N}(\mu^{*} + A_{2}(Y - \mu), B_{2} ).
    \end{equation*}

\textbf{Sampling from the posterior }distribution with constraints has been a challenging task in previous work \citet{riihimaki2010gaussian}. However, \citet{agrell2019gaussian} proposed to use a new method based on simulation via minimax tilting proposed by \citet{botev2017normal}. This sampling approach was proposed for high-dimensional exact sampling and was shown to efficient and fast compared to previous approaches like rejection sampling and Gibb sampling \citet{kotecha1999gibbs}. \\

\textbf{The specification of the location of virtual observations } $Z^v$ can have an important effect on the efficiency and scalability of the monotonic Gaussian process. \citet{agrell2019gaussian} proposed to have a suboptimization problem to find the optimal location. The idea is to iteratively place virtual observation locations where the probability that the constraint holds is low. However, this optimization becomes suboptimal when the dimension of the problem grows. Given that our function is monotonic with respect to only one dimension, we chose to define linearly spaced locations with respect to dimension $t$. The number of points is defined based on the kernel:
\begin{itemize}
    \item ED kernel: The virtual observations locations will mainly enforce the direction of the monotonicity, therefore adding only two locations is sufficient.
    \item RBF Kernel: The number and location of virtual observations depends on the smoothness (lengthscale) of the kernel. The distance between every two virtual observations should be smaller than the lengthscale in order to maintain the monotonicity and avoid any fluctuations. 
\end{itemize}

For more details about the the efficient posterior computation of the monotonic GP we refer the reader to \citet{agrell2019gaussian}.

\subsection*{kernels derivatives}
The computation of the posterior of the monotonic Gaussian process requires the definition of derivatives of the kernel function. In this work we consider monotonicity with respect to one dimension $t$. Therefore, kernel derivatives would be defined as follow  
\begin{align}
    &\frac{\partial}{\partial t} K([\vec{x},t],[\vec{x'},t'])=K_x(x,x')\times \frac{\partial}{\partial t} K_t(t,t')\\
   &\frac{\partial}{\partial t \partial t'} K([\vec{x},t],[\vec{x'},t'])=K_x(x,x')\times \frac{\partial}{\partial t \partial t'} K_t(t,t')
\end{align}
In this work $t$ is a single dimensional variables. However, for sake of generality , we provide the kernel derivatives for the general case where $t$ can be multi-dimensional. We define $d_t$ as the $t$. 
In our experiments, we focus mainly on cases where the kernel over dimension $t$ is an ED kernel. However, Our proposed method is not restrictive. In cases where the learning curve is not exponentially decaying, an RBF kernel with monotonicity over dimension  $t$ can be used. We provide the derivatives for both kernels\\
\textbf{Exponential Decay Kernel}
\begin{align}
&K_t(\vec{t},\vec{t'}) =w+(\frac{\vec{t}}{\boldsymbol{\beta}}+\frac{\vec{t'}}{\boldsymbol{\beta}}+1)^{-\alpha}\\ 
   & \frac{\partial}{\partial t_j'} K_t(\vec{t},\vec{t'}) =- \frac{\alpha}{\beta_j} (\frac{\vec{t}}{\boldsymbol{\beta}}+\frac{\vec{t'}}{\boldsymbol{\beta}}+1)^{-\alpha-1}\\
      & \frac{\partial}{\partial t_j\partial t_j'} K_t(\vec{t},\vec{t'}) =  \frac{\alpha(\alpha+1)}{\beta_j^2} (\frac{\vec{t}}{\boldsymbol{\beta}}+\frac{\vec{t'}}{\boldsymbol{\beta}}+1)^{-\alpha-2}\\
        & \frac{\partial}{\partial t_i\partial t_j'} K_t(\vec{t},\vec{t'}) =  \frac{\alpha(\alpha+1)}{\beta_i\beta_j} (\frac{\vec{t}}{\boldsymbol{\beta}}+\frac{\vec{t'}}{\boldsymbol{\beta}}+1)^{-\alpha-2}
\end{align}
\textbf{Radial basis function Kernel}
\begin{align}
&K_t(\vec{t},\vec{t'}) =exp(\frac{-1}{2} \sum_{i=1}^{d_t} \frac{(t_i-t'_i)^2}{l_i})\\
   & \frac{\partial}{\partial t_j'} K_t(\vec{t},\vec{t'}) =\frac{t_j-t'_j}{l_j^2}K_t(\vec{t},\vec{t'})\\
      & \frac{\partial}{\partial t_j\partial t_j'} K_t(\vec{t},\vec{t'}) = \frac{1}{l_j^2}(1-\frac{t_j-t'_j}{l_j^2}) K_t(\vec{t},\vec{t'})\\
        & \frac{\partial}{\partial t_i\partial t_j'} K_t(\vec{t},\vec{t'}) =  -\frac{t_j-t'_j}{l_j^2} \frac{t_i-t'_i}{l_i^2}K_t(\vec{t},\vec{t'})
\end{align}

\section{Experimental Setup and Additional Results} \label{experiments_details}
\subsection{Experimental Setup Details}\label{setup_details}
\textbf{Logistic Regression with MNIST:}\\
We train the logistic regression classifier on the MNIST image dataset \citet{lecun1998gradient}. The dataset consists of 70,000 images categorized into 10 classes. We use 80\% for training and 10\% for validation. We optimize the model over three hyperparameter: the learning rate $\in [10^{-6},1]$, the $L_2$ regularization $\in [0,1]$ and the batch size $\in [20,2000]$. We apply a log transformation to the learning rate and batch size. We set the maximum number of epochs to $100$.

\textbf{MLP with Olivetti and Covtype:}\\
We train a multi-layer perceptron with two fully connected layers on the Olivetti dataset \citet{samaria1994parameterisation} and Covtype dataset. We use 10\% of the data for the validation set. We optimize four hyperparameters learning rate $\in [10^{-6},1]$, batch size $\in [8,128]$ for Olivetti and $\in [32,1024]$ for Covtype , the $L_2$ regularization $\in [10^{-7},10^{-3}]$ and the momentum $\in [0.1,0.9]$. We apply a log transformation to the learning rate, the batch size and the $L_2$ regularization. We set the maximum number of epochs to $100$. The experiment with Covtype dataset was run on Tesla V100 GPU machine and the experiment on Olivetti was run on a CPU machine with Intel(R) Core(TM) i9-7960X CPU 2.80GHz.

\begin{figure*}[h!]
    \centering
    \begin{minipage}{0.49\textwidth}
        \centering
        \includegraphics[width=1\textwidth]{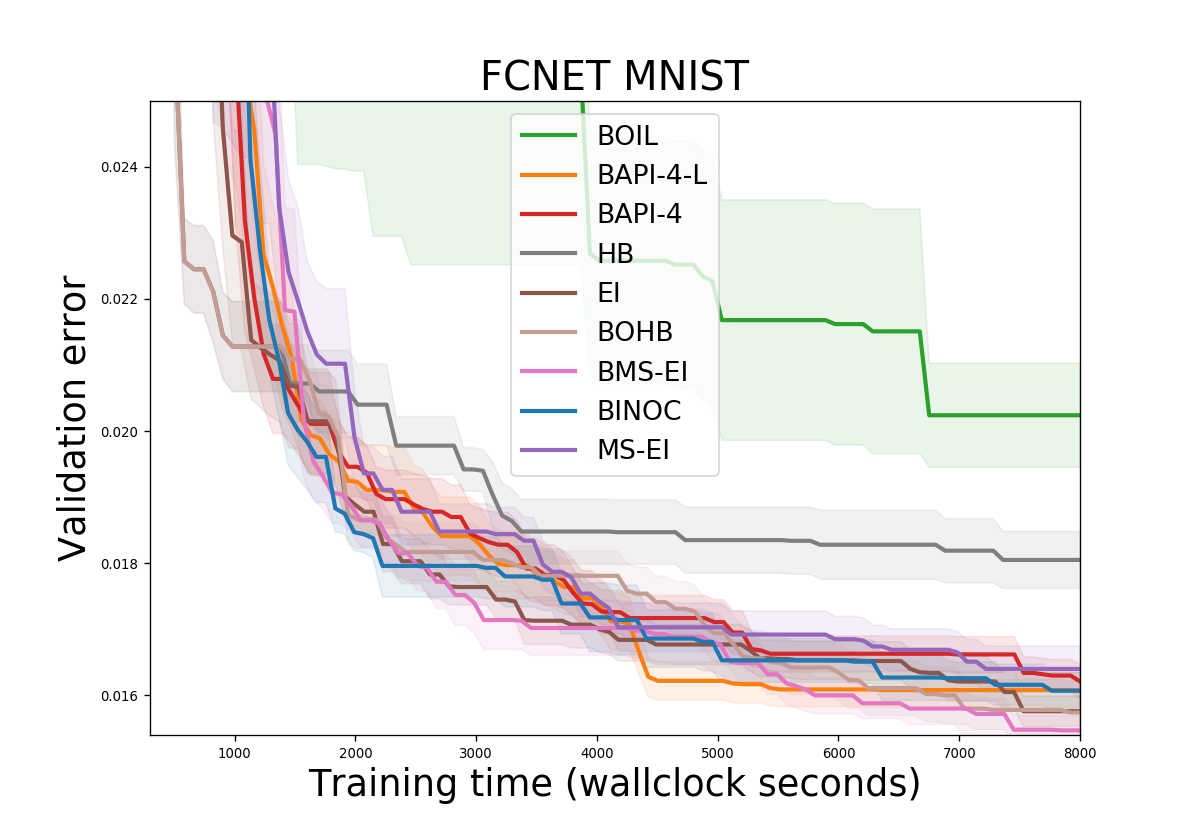} 
    \end{minipage}
    \begin{minipage}{0.49\textwidth}
        \centering
        \includegraphics[width=1\textwidth]{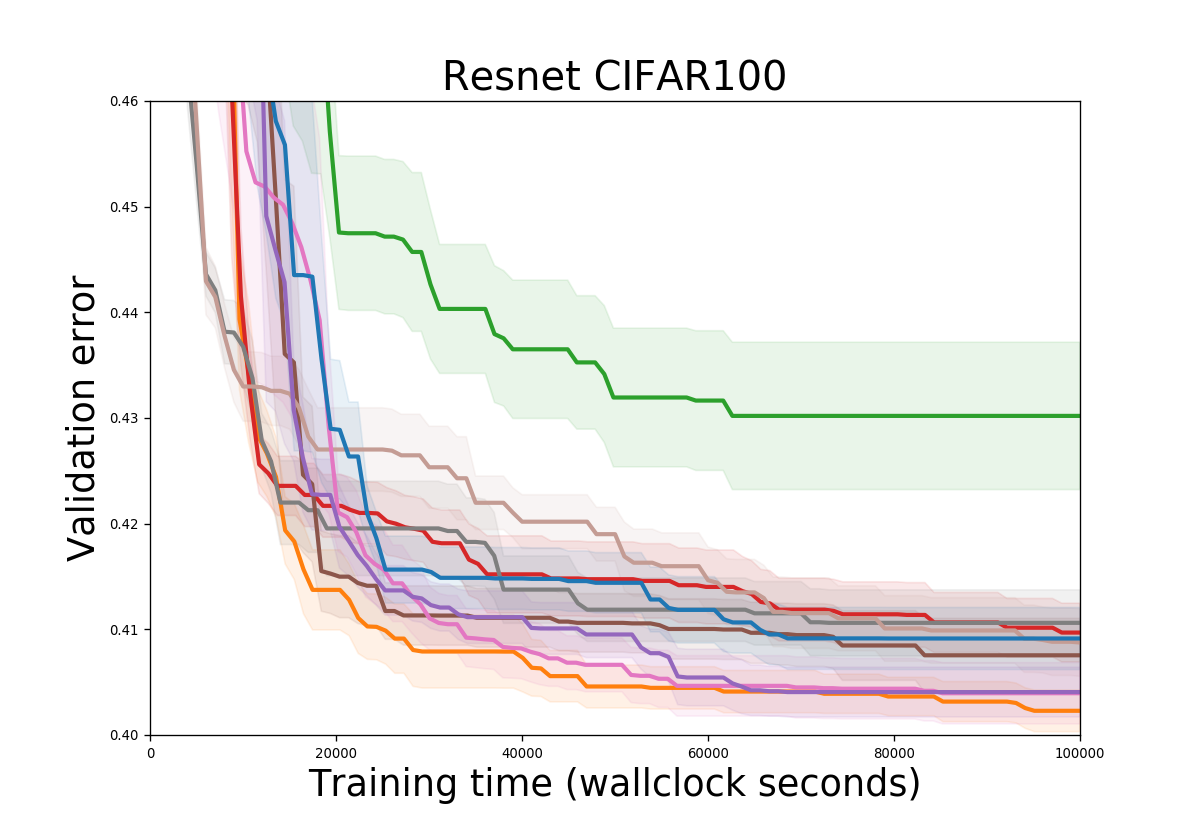} 
    \end{minipage} \hfill
    \begin{minipage}{0.49\textwidth}
        \centering
        \includegraphics[width=1\textwidth]{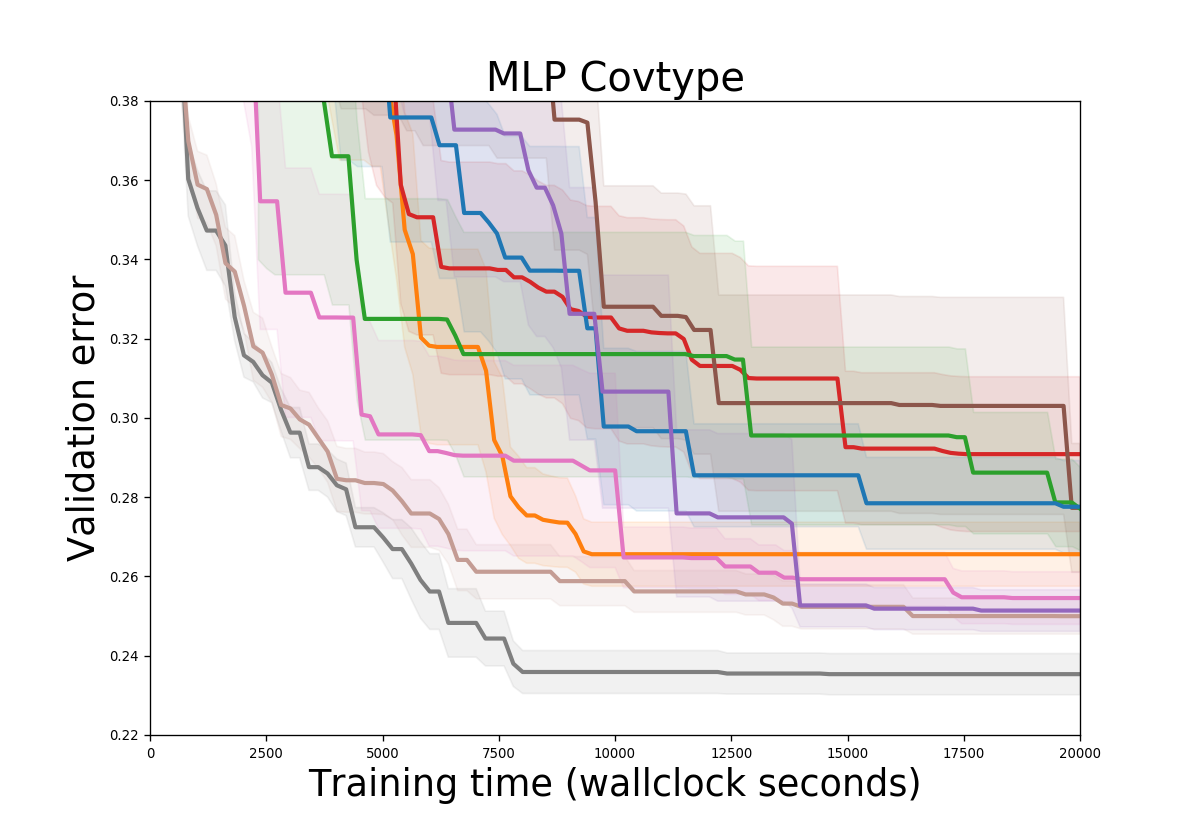} 
    \end{minipage} 
\caption{Results of validation error $\pm$ standard error for different baselines and our proposed approach on ResNet with $t_{max}=100$, FCNET with $t_{max}=25$ and MLP-Covtype with $t_{max}=100$ }
\label{experiments_figure2}
\end{figure*}

\begin{figure*}[h!]
    \centering
    \begin{minipage}{0.47\textwidth}
        \centering
        \includegraphics[width=1\textwidth]{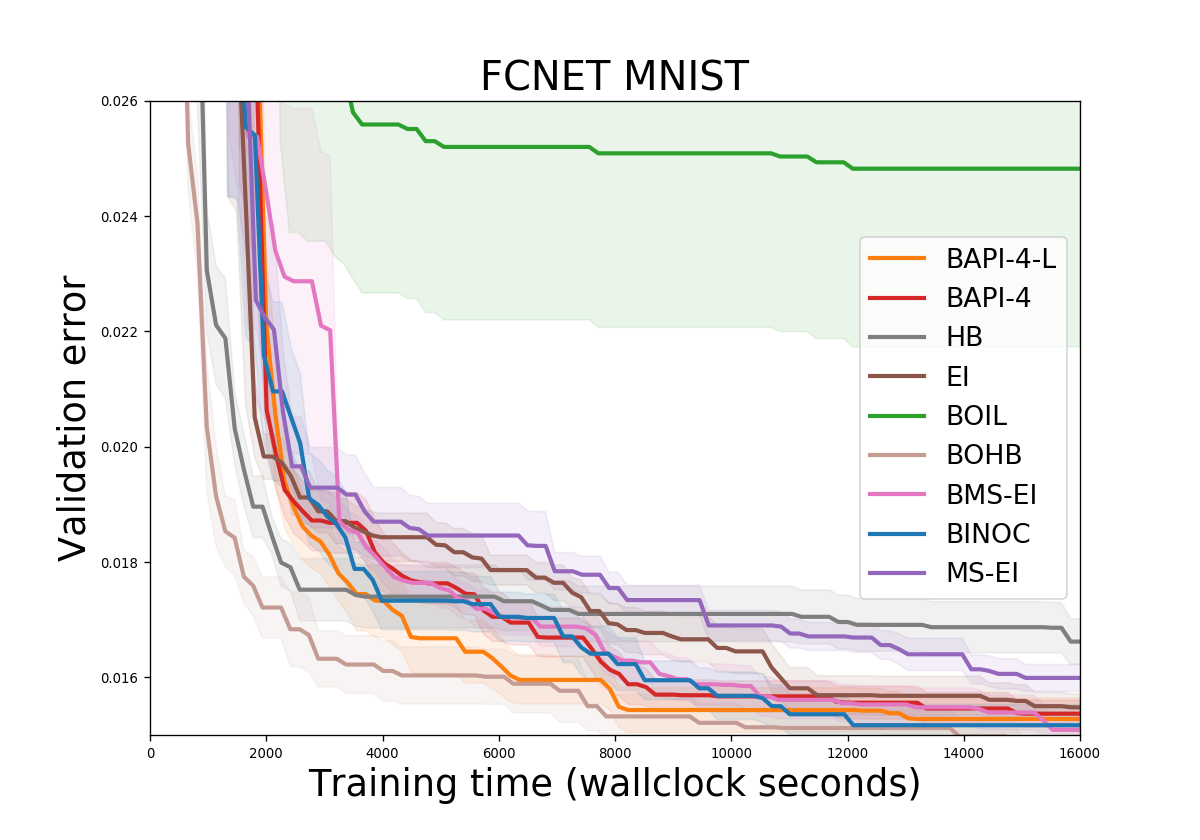} 
    \end{minipage}\hfill
    \begin{minipage}{0.47\textwidth}
        \centering
        \includegraphics[width=1\textwidth]{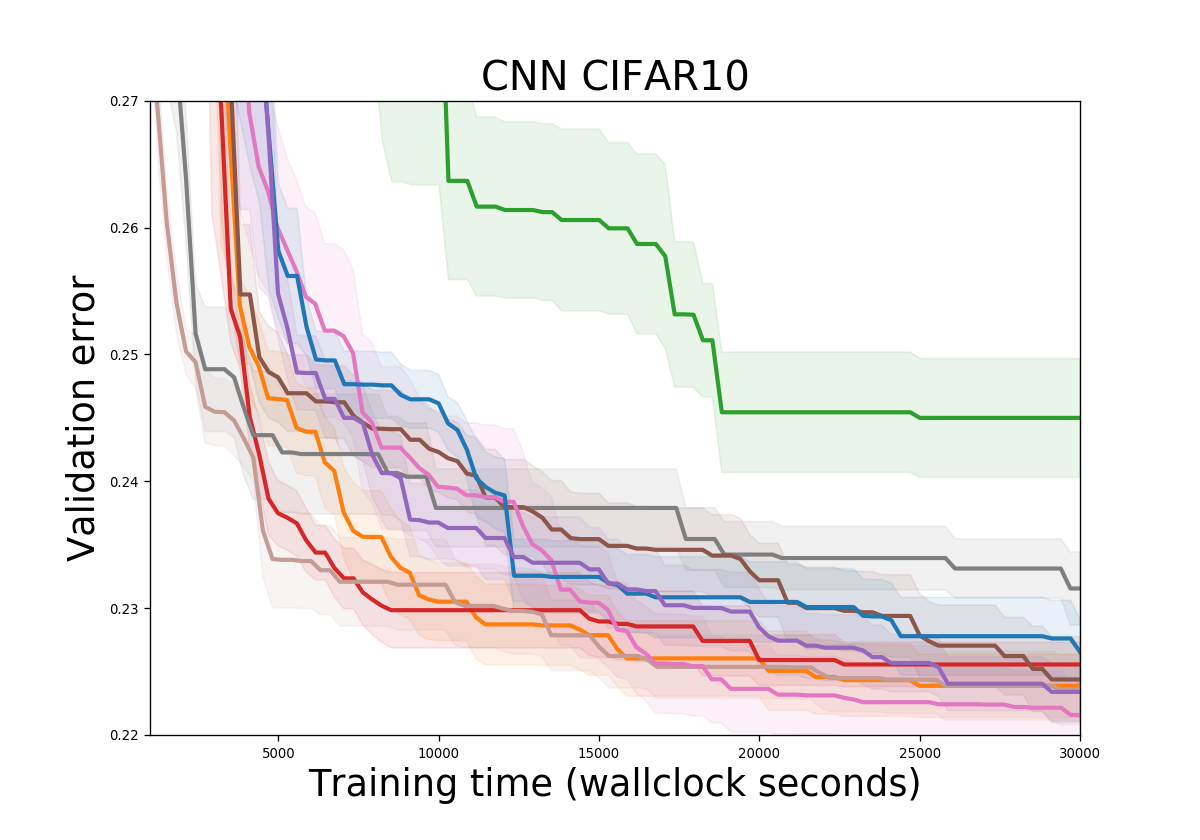} 
    \end{minipage} 
    \begin{minipage}{0.47\textwidth}
        \centering
        \includegraphics[width=1\textwidth]{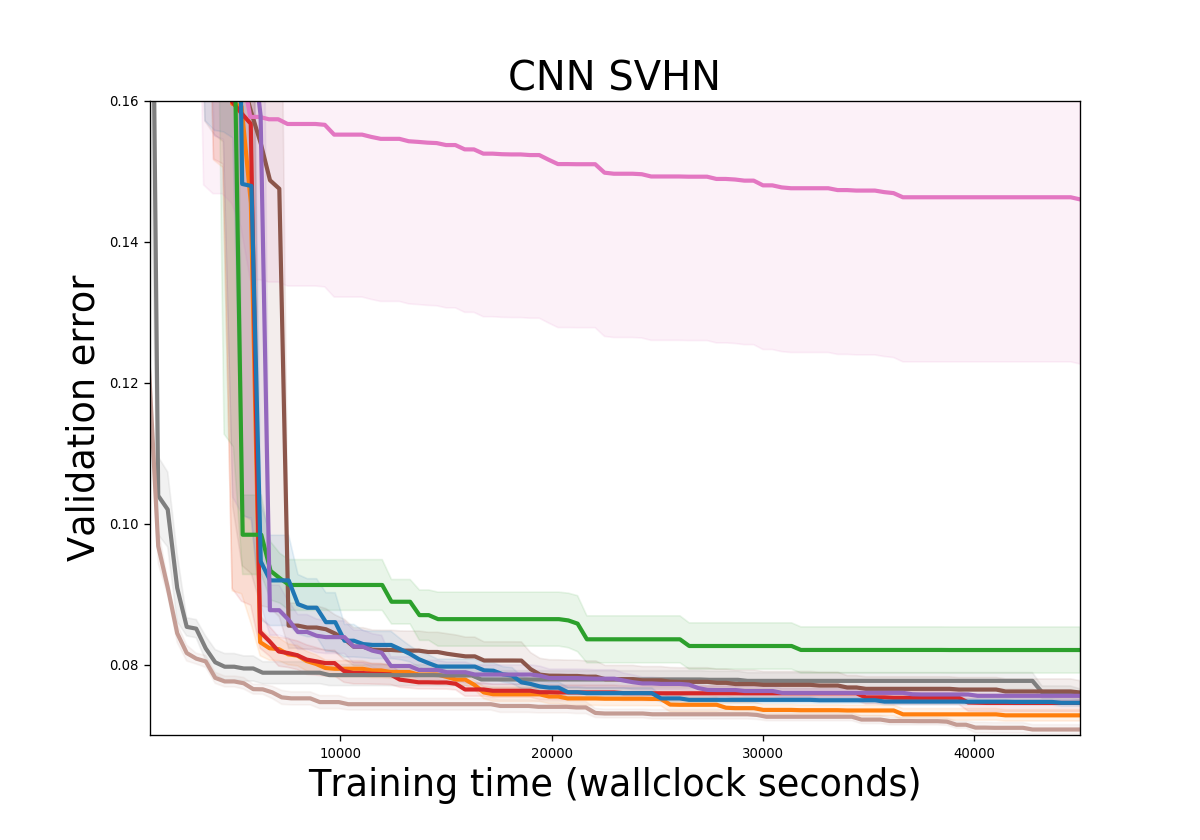} 
    \end{minipage} 
\caption{Results of validation error $\pm$ standard error for different baselines and our proposed approach on FCNET-MNIST, CNN-CIFAR10 and CNN-SVHN with $t_{max}=50$}
\label{experiments_figure3}
\end{figure*}

\textbf{FCNET MNIST:}\\
We train a fully connected network with on the MNIST dataset. We use 50,000 images for the training set and 10,000 images for the validation set. We optimize six hyperparameters learning rate $\in [10^{-6},0.1]$, batch size $\in [32,1024]$, , the $L_2$ regularization $\in [10^{-7},10^{-3}]$, the momentum $\in [0.1,0.9]$, the number of hidden layers $\in [1,4]$ and the size of hidden layers $\in [100,1000]$  We apply a log transformation to the learning rate and the batch size.  We set evaluate all algorithms on two different setups where in figure \ref{experiments_figure2} we report the results with the maximum number of epochs set to $t_{max}=25$ and in figure \ref{experiments_figure3}  we report the results with the maximum number of epochs set to $t_{max}=50$. The total wallclock time budget is extended accordingly. These experiments were run on Tesla V100 GPU machine.

\textbf{CNN with CIFAR10 and SVHN:}\\
We train a CNN model on two image datasets CIFAR10 \citep{krizhevsky2009learning} and the Street View House Numbers (SVHN) \citep{netzer2011reading}.
For CIFAR10 we use 40,000 image for the training set and 10,000 for the validations set. For SVHN 63,257 image for the training set and 10,000 for the validations set.
We optimize six hyperparameters: the batch size$\in [32,1024]$, the learning rate$\in [10^{-6},0.1]$, the momentum$\in [0.1,0.9]$, the $L_2$ regularization$\in [10^{-7},10^{-3}]$, the number of convolutional filters$\in [32,256]$, and the number of dense units $\in [64,512]$. We apply a log transformation to the learning rate, the batch size. We set evaluate all algorithms on two different setups where in figure \ref{experiments_figure} we report the results with the maximum number of epochs set to $t_{max}=25$ and in figure \ref{experiments_figure3}  we report the results with the maximum number of epochs set to $t_{max}=50$. The total wallclock time budget is extended accordingly. These experiments were run on Tesla P100 GPU machine.

\textbf{Resnet with CIFAR100:}\\
We train a ResNet model on a the image dataset CIFAR100 \citep{krizhevsky2009learning}. We employ 40,000 images for the training set and 10,000 for the validations set. We optimize six hyperparameters: the batch size $ \in [32,512]$, the learning rate $ \in [1e-6,1e-1]$, the momentum $\in [0.1,0.9]$, the $L_2$ regularization$\in [ 1e-7, 1e-3]$, the number of convolutional filters$\in [32,256]$, and the number of layers $\in [10,18]$. We report the results with the maximum number of epochs set to $t_{max}=100$ in Figure \ref{experiments_figure2}. The total wall-clock time budget is extended accordingly. These experiments were run on a Tesla V100 GPU machine. 

\textbf{DQN CartPole:}\\
We train a Dueling DQN (DDQN) \citep{wang2016dueling}
agent in the CartPole-v0 environment. We employ the same setting proposed by \citet{nguyen2020bayesian}. We optimize two hyperparameters: the discount factor $\in [0.8, 1]$ and the
learning rate for the model $\in [1e-6, 0.01]$. We vary the number of episodes from 200 to 500. We map the episodes into epochs with each three episodes equivalent to one epoch, resulting in a maximum number of epochs $t_{max}=100$. We report the results in Figure \ref{experiments_figure3}. The total wall-clock time budget is extended accordingly. These experiments were run on a 1 core of a Xeon CPU machine. 

\textbf{A2C Reacher:}\\
We train a Advantage Actor Critic (A2C) \citep{mnih2016asynchronous}
agent in the Reacher-v2 environment. We employ the same setting proposed by \citet{nguyen2020bayesian}. We optimize three hyperparameters: the discount factor $\in [0.8, 1]$, the
learning rate for the actor $\in [1e-6, 0.01]$, and the
learning rate for the critic $\in [1e-6, 0.01]$. We vary the number of episodes from 200 to 500. We map the episodes into epochs with each three episodes equivalent to one epoch, resulting in a maximum number of epochs $t_{max}=100$. We report the results in Figure \ref{experiments_figure3}. The total wall-clock time budget is extended accordingly. These experiments were run on a 1 core of a Xeon CPU machine.  

\textbf{A2C Inverted Pendulum:}\\
We train a Advantage Actor Critic (A2C) \citep{mnih2016asynchronous}
agent in the InvertedPendulum-v2 environment. We employ the same setting proposed by \citet{nguyen2020bayesian}. We optimize three hyperparameters: the discount factor $\in [0.8, 1]$, the
learning rate for the actor $\in [1e-6, 0.01]$, and the
learning rate for the critic $\in [1e-6, 0.01]$. We vary the number of episodes from 700 to 1500. We map the episodes into epochs with each eight episodes equivalent to one epoch, resulting in a maximum number of epochs $t_{max}=100$. We report the results in Figure \ref{experiments_figure3}. The total wall-clock time budget is extended accordingly. These experiments were run on a 1 core of a Xeon CPU machine.

\subsection{Additional Results and Discussion}\label{additiona_results}
We report additional results of our BAPI approach and existing baselines. We report an additional variant of our algorithm that we name BAPI-4-L. We test the case where we do not add additional points from each curve but rather use only the last epoch. We notice that this variant performs competitively and sometimes better than adding additional points to the curve. This opens a discussion about the utility of leveraging additional data points from each curve especially while using the monotonic GP. It is important to note that previous methods built for HPO frequently suggest using additional points. This includes approaches proposed in the papers by \citet{nguyen2020bayesian,dai2019bayesian}, and \citet{wu2020practical}. We plan to work on investigating this problem further to develop a sound theoretical understanding of this phenomenon.

In Figure \ref{experiments_figure3}, we test all algorithms on settings with extended budget and a higher number of maximum epochs $t_{max}=50$. We observe that given a sufficiently large budget, most of the baselines converge to statistically comparable results. We notice that HB, in most of the experiments, is able to reduce the validation error in the beginning but does not always converge to good results. However, BOHB performance was remarkably stronger with a higher number of maximum epochs. Increasing the maximum number of epoch enables BOHB to evaluate a larger number of configurations at a low budget and therefore we can see a significant drop in the validation error earlier than all baselines. The results in Figure \ref{experiments_figure2} and Figure \ref{experiments_figure3} show that BAPI-4 becomes less competitive
when the total budget is significantly increased and the maximum number of epochs is higher, but also provides results suggesting that performance loss can be brought back by adjusting pruning behavior in BAPI-4-L.

We report additional results for reinforcement learning experiments optimized with RBF kernel over the number of epochs. Figure \ref{experiments_figure3} shows the increasing discounted cumulative reward with discount factor 0.9 as suggested by \cite{dai2019bayesian}. The results show that BAPI-4 performs better or similar to the baselines. We observe that HB and BOHB performance degrades significantly with RL experiments most likely because they do not account for the possibility that the learning curve can be flat in the middle. We additionally notice that BAPI-4-L performance is competitive but worse than BAPI-4. One candidate reason for this behavior is due to the use of the RBF kernel, where adding intermediate points from the curve can be more crucial to avoid fluctuations.   

\subsection{Ablation: GP without enforced monotonicity}
We provide an ablation study where we run our algorithm using a GP without enforced monotonicity to show the benefit of using a monotonic GP. The first figure illustrates differences in learning curve extrapolation between a \textbf{GP with enforced monotonicity and vanilla RBF GP}. The RBF GP fluctuates further from evaluated points rendering extrapolation highly uncertain (as well as inaccurate).
Basing an estimation of optimal stopping time on this vanilla model directly affects budgeted HP optimization performance as shown in the \textbf{ablation study} displayed in the second and third figures. 
These show the performance of BAPI with a Non-Monotonic GP (BAPI-4-NM) is inferior to BAPI with monotonic GP. However, BAPI-4-NM shows competitive performance that might be associated with its budget-aware planning strategy.

\begin{figure}[h!]
    \centering
    \begin{minipage}{0.5\textwidth}
        \centering
        \includegraphics[width=0.7\linewidth]{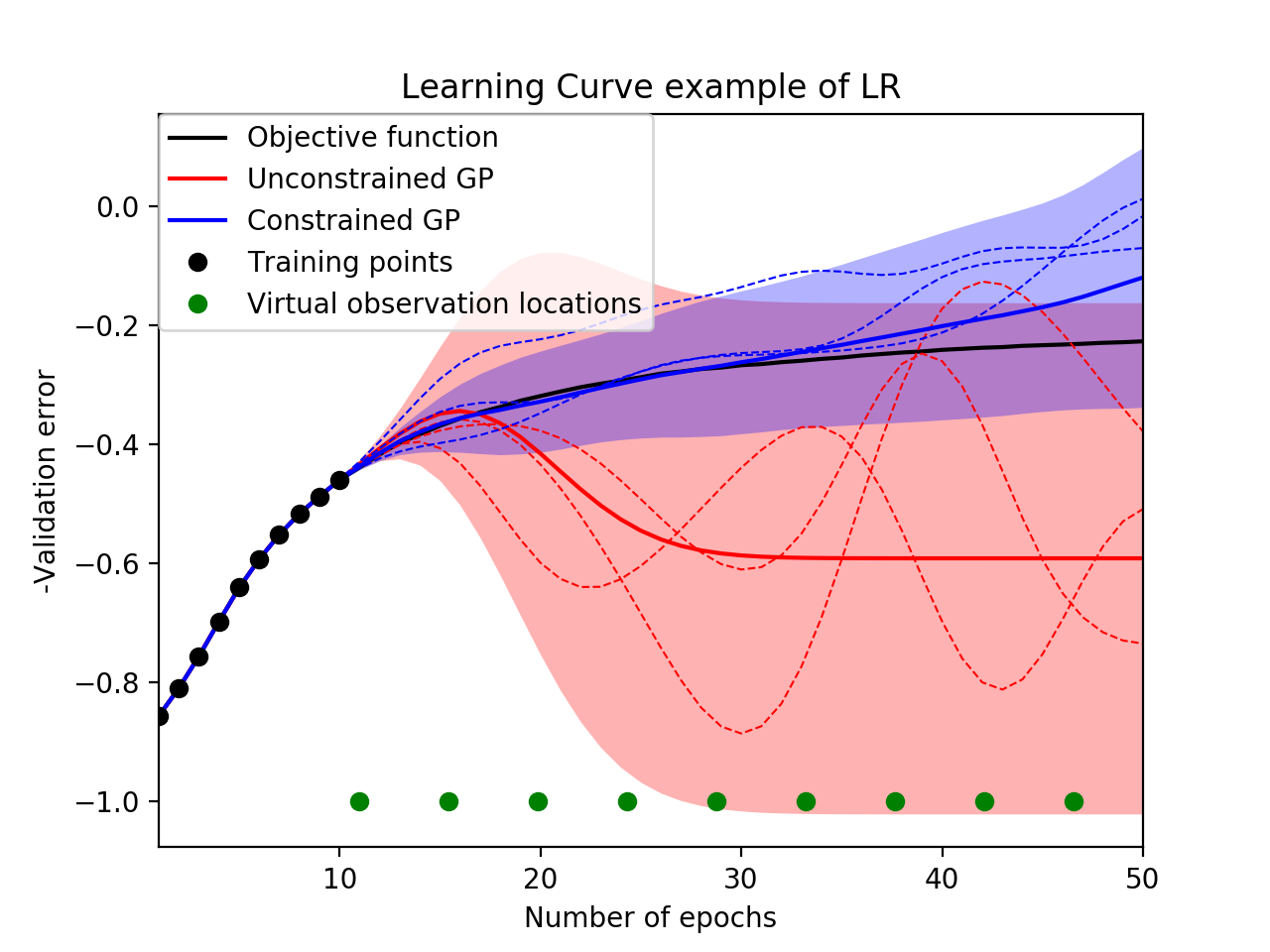}
    \end{minipage}
    \begin{minipage}{0.23\textwidth}
        \centering
        \includegraphics[width=1\textwidth]{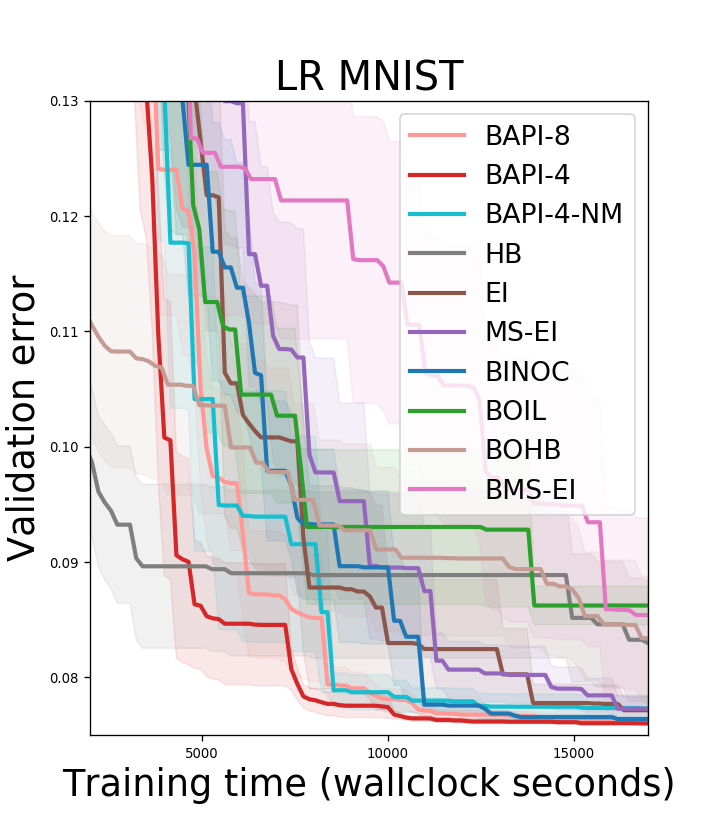} 
    \end{minipage}
    \begin{minipage}{0.23\textwidth}
        \centering
        \includegraphics[width=1\textwidth]{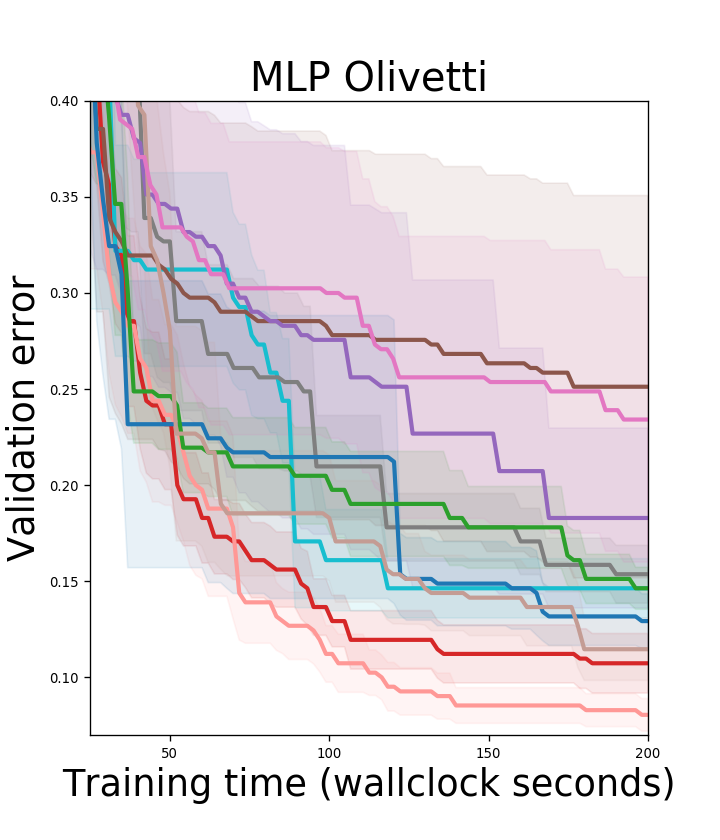} 
    \end{minipage} 
\label{ablation_figure}
\end{figure}
\begin{figure*}[h!]
    \centering
    \begin{minipage}{0.49\textwidth}
        \centering
        \includegraphics[width=1\textwidth]{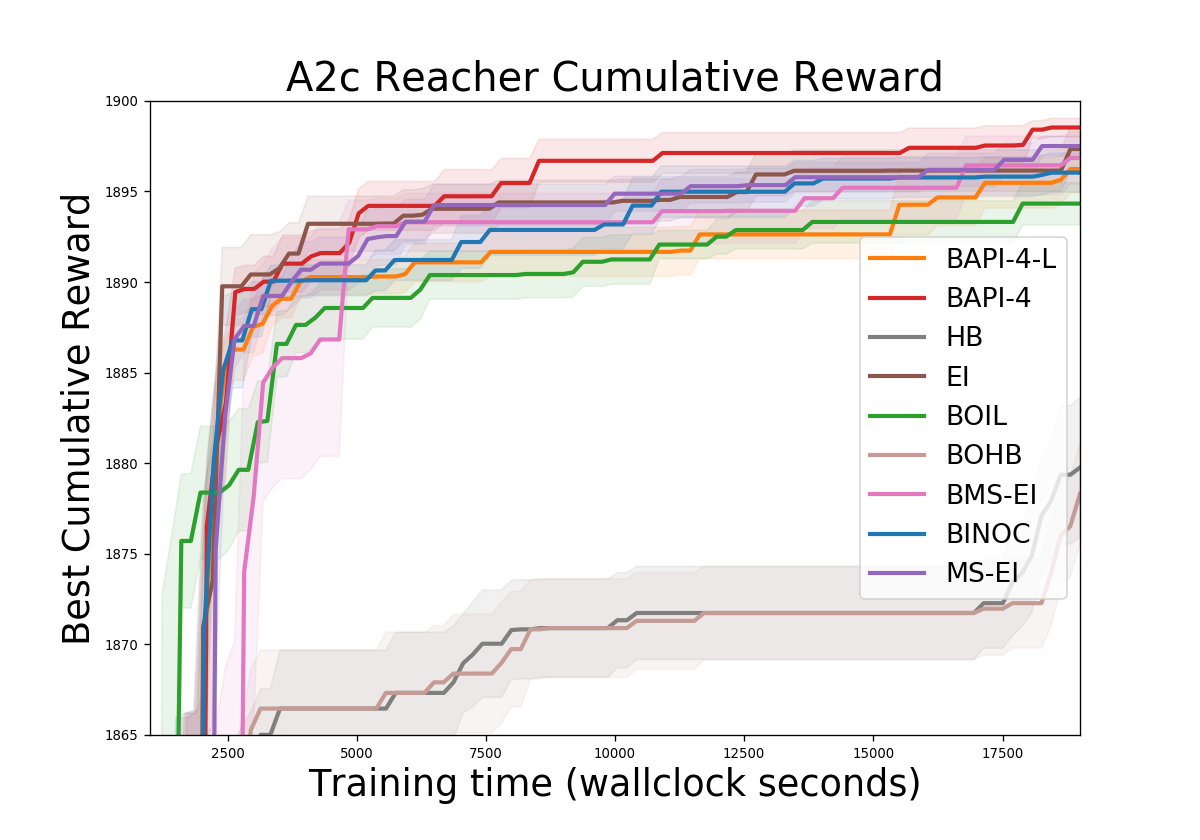} 
    \end{minipage}\hfill
    \begin{minipage}{0.49\textwidth}
        \centering
        \includegraphics[width=1\textwidth]{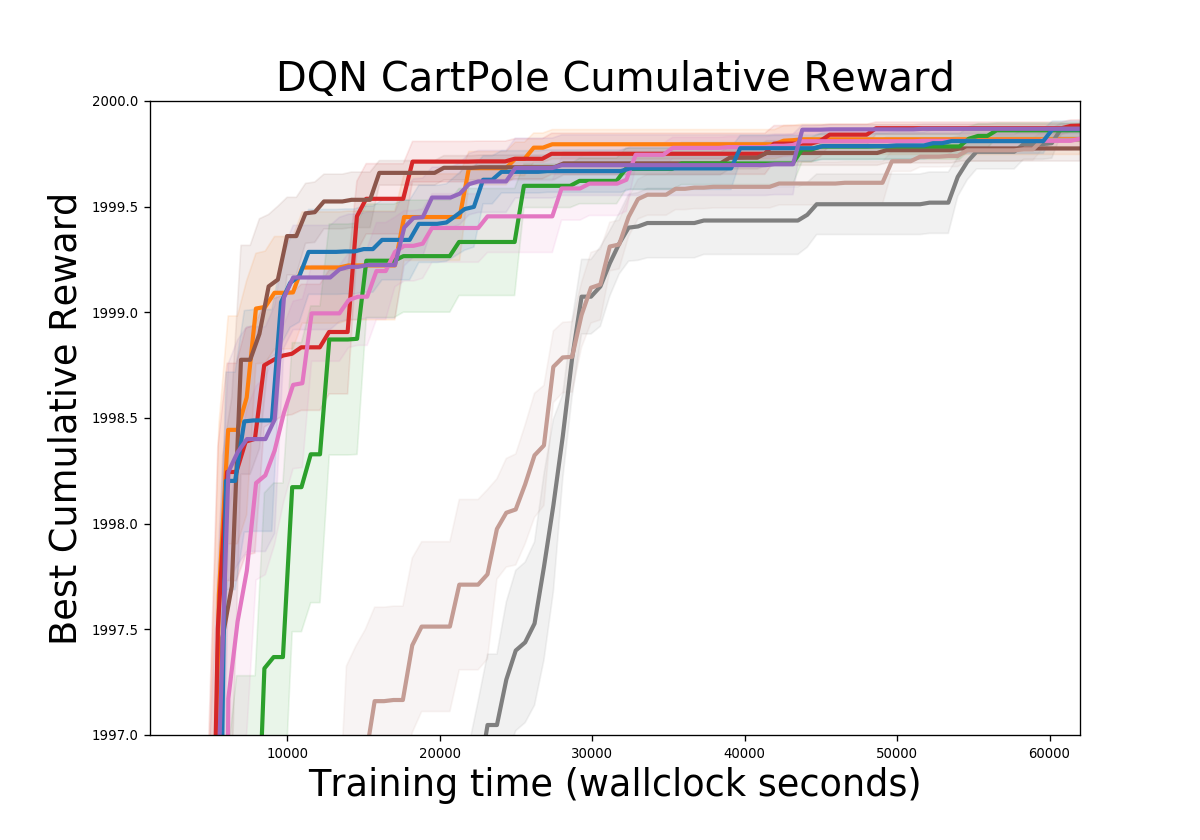} 
    \end{minipage} 
    \begin{minipage}{0.49\textwidth}
        \centering
        \includegraphics[width=1\textwidth]{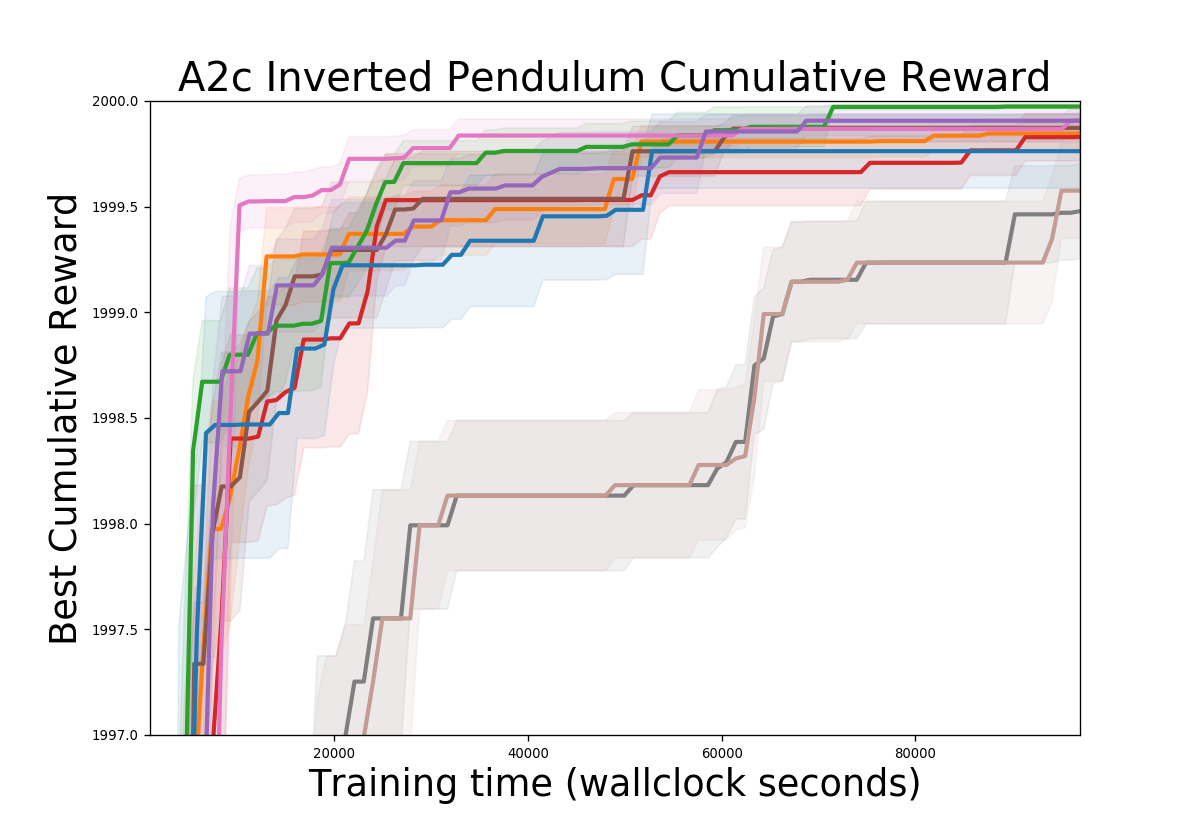} 
    \end{minipage} 
\caption{Results of Cumulative discounted reward $\pm$ standard error for different baselines and our proposed approach on A2C Reacher, DQN Cartpole, and A2C Inverted Pendulum}

\label{experiments_figure4}
\end{figure*}